%% file: main.tex
\PassOptionsToPackage{sort}{natbib} 

\documentclass{article} 
\usepackage{iclr2027_conference,times}

\input{math_commands.tex}

\definecolor{cvprblue}{rgb}{0.21,0.49,0.74}
\usepackage[colorlinks,citecolor=cvprblue,citecolor=cvprblue,linkcolor=cvprblue,urlcolor=cvprblue]{hyperref}

\usepackage{url}
\usepackage{booktabs}
\usepackage{graphicx}
\usepackage{multirow}
\usepackage[dvipsnames]{xcolor}
\usepackage{colortbl}
\definecolor{tabfirst}{rgb}{1, 0.7, 0.7}      
\definecolor{tabsecond}{rgb}{1, 0.85, 0.7}    
\definecolor{tabthird}{rgb}{0.98, 0.96, 0.7}  
\newcommand{\firstbox}[1]{{\setlength{\fboxsep}{1pt}\colorbox{tabfirst}{#1}}}
\newcommand{\secondbox}[1]{{\setlength{\fboxsep}{1pt}\colorbox{tabsecond}{#1}}}
\newcommand{\thirdbox}[1]{{\setlength{\fboxsep}{1pt}\colorbox{tabthird}{#1}}}
\usepackage{lipsum}
\usepackage{pifont}
\usepackage{wrapfig}
\usepackage{etoc}
\usepackage{float}

\usepackage{enumitem}
\usepackage[font=small]{caption}
\usepackage[capitalize]{cleveref}
\crefname{section}{Sec.}{Section}
\usepackage{tabularx}
\usepackage{makecell}
\usepackage{etoolbox,siunitx}
\robustify\bfseries
\definecolor{mygreen}{RGB}{0,143,0}
\definecolor{ppgA}{HTML}{BD6125}
\definecolor{ppgB}{HTML}{F4A841}
\usepackage{xstring,pgffor}
\newcommand{\gradtext}[1]{%
    \StrLen{#1}[\strlen]%
    \foreach \i in {1,...,\strlen} {%
        \StrChar{#1}{\i}[\thechar]%
        \pgfmathtruncatemacro{\pct}{100 - (\i-1)*100/(\strlen-1)}%
        \textcolor{ppgA!\pct!ppgB}{\thechar}%
    }%
}

\newcommand{\cmark}{\ding{51}}
\newcommand{\xmark}{\ding{55}}
\newcommand{\paragrapht}[1]{\noindent\textbf{#1}}

\newcommand{\ours}{FL4N}

\newcommand{\aftertab}{\vspace{-0.3em}}
\newcommand{\aftertabcaption}{\vspace{-0.3em}}
\newcommand{\afterfig}{\vspace{-0.3em}}
\newcommand{\beforetfigcaption}{\vspace{-0.1em}}

\title{\ours: Feed-forward Latent Model for\\ 4D Novel View Synthesis}

\author{%
\textbf{Eunbeen Hong}$^{1}$ \quad
\textbf{Hyeonseo Yu}$^{1}$ \quad
\textbf{Honggyu An}$^{1}$ \quad
\textbf{Jooho Son}$^{1}$ \quad
\textbf{Hyunah Ko}$^{1}$ \\
\textbf{Soowon Son}$^{1}$ \quad
\textbf{Chaehyun Kim}$^{1}$ \quad
\textbf{Jahyeok Koo}$^{1}$ \quad
\textbf{Jisoo Kim}$^{1}$ \quad
\textbf{Junhwa Hur}$^{2\dagger}$ \\ 
\textbf{Seungryong Kim}$^{1\dagger}$
\\[6pt]
{\normalfont $^{1}$KAIST AI \qquad $^{2}$Google DeepMind}
\\[2pt]
{\normalfont\fontfamily{cmtt}\selectfont\href{https://cvlab-kaist.github.io/FL4N/}{\gradtext{https://cvlab-kaist.github.io/FL4N/}}}
}

\iclrfinalcopy
\begin{document}

\maketitle
\lhead{\ours: Feed-forward Latent Model for 4D Novel View Synthesis}
{\renewcommand{\thefootnote}{\fnsymbol{footnote}}\footnotetext[2]{Corresponding authors.}}
\etocsettocdepth.toc{none}

\input{figures/concept}

\input{sections/0_abstraction}

\input{sections/1_intro}

\input{sections/2_relworks}

\input{sections/3_method}
\input{sections/4_experiments}

\input{sections/5_conclusion}

\bibliography{main}
\bibliographystyle{iclr2027_conference}

\clearpage

\input{sections/Appendix}

\end{document}

%% file: math_commands.tex
\usepackage{amsmath,amsfonts,bm,xspace}

\newcommand{\figleft}{{\em (Left)}}

\newcommand{\figright}{{\em (Right)}}

\def\eqref#1{equation~\ref{#1}}

\def\1{\bm{1}}

\DeclareMathAlphabet{\mathsfit}{\encodingdefault}{\sfdefault}{m}{sl}
\SetMathAlphabet{\mathsfit}{bold}{\encodingdefault}{\sfdefault}{bx}{n}

\providecommand*{\onedot}{\futurelet\lettoken\onedothelper}
\providecommand*{\onedothelper}{\ifx\lettoken.\else.\null\fi\xspace}

\providecommand*{\eg}{\emph{e.\thinspace{}g}\onedot}

\providecommand*{\ie}{\emph{i.\thinspace{}e}\onedot}

%% file: figures/concept.tex
\begin{figure}[H]
\centering

\includegraphics[width=1.0\textwidth]{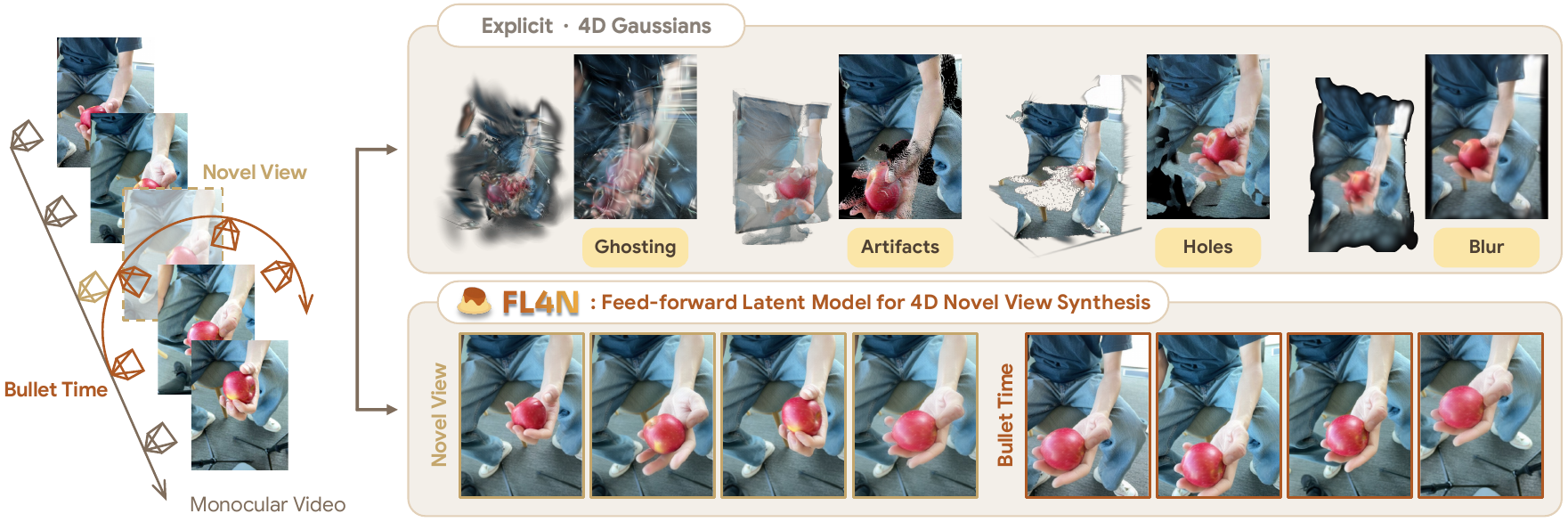}

\beforetfigcaption
\caption{\textbf{Teaser.} \textbf{\ours{}} is the first feed-forward latent 4D novel view synthesis model. It models dynamics in a latent space to render temporally coherent frames, whereas Gaussian-based methods such as 4DGT~\citep{4dgt}, MoVieS~\citep{lin2026movies}, NeoVerse~\citep{yang2026neoverse}, and C4G~\citep{c4g} suffer from temporal ghosting, artifacts, holes, or blur.}
\label{fig:teaser}

\vspace{-.3em}
\afterfig

\end{figure}

%% file: sections/0_abstraction.tex
\begin{abstract}

4D novel view synthesis (NVS) aims to render dynamic scenes across viewpoints and timestamps, which requires jointly inferring 3D structure and temporal dynamics.
Existing monocular feed-forward methods often rely on explicit 4D Gaussian representations, which are prone to temporal inconsistency and severe rendering artifacts.
We present \textbf{FL4N}, the first feed-forward latent 4D NVS model.
Instead of relying on explicit 4D primitives, FL4N encodes a 4D scene prior into a unified latent representation across space and time.
Our core insight is that learning temporal dynamics as residuals on top of a pretrained static latent prior enables effective 4D modeling with limited dynamic training data.
To this end, we introduce a lightweight dynamic encoder that leverages static scene features to disentangle object motion from camera-induced parallax.
Trained solely with an image synthesis objective, FL4N achieves state-of-the-art 4D NVS quality while rendering in real time.
Its learned latent representation generalizes to real-world videos and disentangles viewpoint from time.

\end{abstract}

%% file: sections/1_intro.tex
\section{Introduction}
\label{sec:intro}

4D novel view synthesis (NVS) is a cornerstone of understanding and simulating our ever-changing physical world, enabling broad applications, \eg, immersive graphics~\citep{li20254k4dgen}, world models~\citep{zhu2025aether}, robotics~\citep{kim2026pri4r}, autonomous driving~\citep{song2025coda}, and 4D-aware vision-language models~\citep{cho20264dpqa}.
Unlike static NVS~\citep{lvsm,hong2024pf3plat}, 4D NVS requires capturing both persistent 3D structure and temporal dynamics.
While slow optimization-based methods~\citep{pumarola2021dnerf,wang2025shapeofmotion} and multi-view methods~\citep{xu20244k4d, nopo4d} are impractical in the wild, monocular feed-forward approaches offer a practical, real-time solution by inferring a 4D scene from a casual video. 

However, monocular feed-forward 4D NVS is highly ill-posed, as it must jointly infer entangled 3D structure and object motion from monocular observations alone \citep{wang2025shapeofmotion}.
This requires the model to learn a strong 4D prior. 
To make this tractable, existing methods \citep{4dgt,lin2026movies,yang2026neoverse} rely on the explicit 3D inductive bias of 3D Gaussians \citep{kerbl20233d} with added 3D motion and dense geometry and motion supervision.
However, this reliance on explicit 3D primitives often bottlenecks 4D rendering quality.
Inaccurate prediction of Gaussian attributes across timestamps often yields rendering artifacts, \eg, temporal ghosting, floaters, and holes \citep{c4g,an2025c3g}, as in \cref{fig:teaser}.
Moreover, time-varying effects (\eg, shadows and illumination changes) must be encoded in per-primitive attributes \citep{li2024spacetimeGS}, and regions with
insufficient primitive coverage tend to be over-smoothed \citep{li2024spacetimeGS, c4g}.

In recent static NVS literature, latent-based methods~\citep{lvsm, szymanowicz2026lagernvs, xfactor} avoid these limitations by learning view synthesis end-to-end with transformers, mapping input views directly to target views conditioned on a target camera, without an explicit 3D representation. 
With minimal 3D inductive biases and large-scale training, these latent models outperform 3D-Gaussian-based methods in visual fidelity~\citep{lvsm,kim2026scaling}. 
However, directly extending the latent approach to dynamic scenes faces a fundamental challenge.
Without explicit 3D priors, learning to disentangle 3D structure and temporal dynamics purely
from data would require massive multi-view dynamic scene datasets~\citep{van2024generativekub4d, bai2025recammaster}, which are much scarcer than static datasets~\citep{watson2025controlling4DiM}.

To resolve this challenge, we propose \textbf{\ours{}}, a \textbf{F}eed-forward \textbf{L}atent-based \textbf{4}D \textbf{N}VS model, which augments a frozen pretrained static latent NVS model with a trainable {dynamic encoder}.
Our key insight is that a 4D latent model does not need to learn dynamic scene representation from scratch: pretrained static latent NVS models already possess a strong, generalizable 3D prior, allowing the dynamic encoder to exclusively focus on modeling temporal changes (\eg, object motion or appearance variation). 
Consequently, our method can be trained with a small amount of dynamic data while fully inheriting the pretrained model's ability to synthesize coherent novel views.

Concretely, we attach the lightweight trainable dynamic encoder alongside the frozen static encoder, similar to ControlNet~\citep{zhang2023adding}. 
The dynamic encoder takes the same input views with a learnable per-view time token and cross-attends to frozen static encoder features to isolate true scene dynamics from camera-induced motion.
A lightweight time module then modulates the residual latent for the queried timestamp, which is residually added to the static latent~\citep{he2016deep} and decoded by the frozen renderer at the target camera pose.
Multi-view videos naturally disentangle viewpoint and temporal variation through same-time views and fixed views across time. Sampling unseen view--time pairs, \ours{} learns this disentanglement through image synthesis with binary dynamic mask supervision, without dense geometry or motion supervision.

We evaluate our approach on synthetic and real datasets under a systematic protocol. To effectively demonstrate the performance of 4D novel view synthesis and the capability of view-time disentanglement, we separate the setting with novel view synthesis, cross-time view synthesis, and novel time synthesis. \ours{} achieves state-of-the-art or competitive performance across datasets in the pose-free setting without test-time pose optimization, while optionally incorporating context camera poses further enables state-of-the-art performance across all three settings.

Our main contributions are summarized as follows:
\begin{itemize}[leftmargin=1em,itemsep=1pt,topsep=1pt]
\item \textbf{First 4D latent NVS model.} We present to jointly model 3D structure and temporal variation in a latent space, rendering novel views at novel timestamps without explicit 3D representations or target-view/time observations.
\item \textbf{A residual formulation for efficient dynamic adaptation.} We formulate 4D NVS as learning a temporal residual on a frozen static representation. This leverages the pretrained scene and viewpoint priors while concentrating learning on temporal variation through a lightweight branch.
\item \textbf{State-of-the-art performance.} We comprehensively benchmark 4D NVS across novel view, cross-time, and novel time synthesis on eight synthetic and real-world benchmarks, achieving state-of-the-art rendering quality on all three settings, with the fastest runtime.
\end{itemize}

\clearpage

%% file: sections/2_relworks.tex
\section{Related work}
\label{sec:related}

\paragrapht{Feed-forward Gaussian splatting for 4D novel view synthesis.}
Explicit feed-forward 4D methods typically represent scenes with 3D Gaussians extended with temporal dynamics, such as velocity~\citep{4dgt, yang2026neoverse} or deformation fields~\citep{lin2026movies,lin2026dgs}.
However, independently modeling motion per Gaussian often disrupts object rigidity and temporal consistency due to incoherent motion among nearby Gaussians~\citep{c4g}, yielding artifacts such as temporal ghosting.
Anchoring Gaussian prediction to each target timestamp~\citep{btimer} can sidestep the issue but is susceptible to view bias by primarily relying on nearby frames while underutilizing observation from distant frames. 
Query-based Gaussian representation~\citep{c4g} alleviates this via dense motion supervision, yet its limited number of queries constrains rendering quality due to insufficient Gaussian coverage, particularly for large outdoor scenes.
Multi-view approaches~\citep{nopo4d} can exploit observations from multiple time-synchronized static cameras, but the setup is far more restrictive than a monocular moving camera.
We introduce a feed-forward latent-based method with minimal geometric inductive bias in a practical but more challenging monocular camera setup.

\paragrapht{Latent-based novel view synthesis.}
Latent-based NVS models synthesize novel views with transformers conditioned on the target pose~\citep{sajjadi2022scene,jiang2025rayzer,lvsm,szymanowicz2026lagernvs,kim2026scaling} or latent pose variable~\citep{xfactor}, bypassing explicit 3D representations.
Although RayDer~\citep{prestel2026rayder} trains on dynamic videos, it still targets static NVS while suppressing object motion via per-view nuisance token that is dropped at inference.
Despite their strong generalization, these methods either assume static scenes or treat dynamic content as variation to be ignored rather than modeled.
Toward dynamic latent representations, DyST~\citep{seitzer2024dyst} extends RUST~\citep{sajjadi2023rust} with a dynamics estimator, yet keeps its scene latent time-invariant.
Temporal variation is decoded via control codes extracted from target-pose and target-time frames, which limits explicit view and time control and risks appearance leakage from these frames.
In contrast, our model constructs a time-conditioned scene latent that is rendered from arbitrary
target cameras, enabling synthesis at novel views and novel timestamps.

\paragrapht{Generative novel view synthesis.}
NVS becomes inherently ambiguous when target views extend into regions unseen in the input views. Deterministic methods maintain strong cross-view consistency on observed content but tend to blur unobserved regions~\citep{prestel2026rayder,szymanowicz2026lagernvs}.
Generative approaches~\citep{van2024generativekub4d, bai2025recammaster, yu2025trajectorycrafter, ren2025gen3c, chen2026reconstruct} can extrapolate plausible content beyond the observed regions, yet their stochastic sampling often compromises cross-view consistency and scene faithfulness~\citep{lvsm, c4g}.
Although generative refinement can post-process rendered views from Gaussian representations~\citep{yang2026neoverse, c4g} or serve as a diffusion decoder for latent-based model~\citep{szymanowicz2026lagernvs}, our work focuses on deterministic NVS and remains orthogonal to such generative refinement directions.

%% file: sections/3_method.tex
\section{Method}
\label{sec:method}

We propose \textbf{\ours{}}, a \textbf{F}eed-forward \textbf{L}atent-based \textbf{4}D \textbf{N}VS model.
Given an $N$-frame monocular video $\{\mathbf{I}_k\}_{k=1}^{N}$, our method synthesizes views from an arbitrary query view $\pi_\text{tgt}$ at target timestamp $t_\text{tgt}$ drawn from the observed time range.
\Cref{fig:arch} shows an overview of \ours{}.
We formulate 4D synthesis as \textit{a residual adaptation} on top of a frozen pretrained static NVS model.
Given the static encoder and the renderer from the pretrained model, we attach a lightweight dynamic encoder that focuses its capacity on isolating dynamic object motion.

In the following, we first review the latent static NVS baseline that forms a basis of our method (\S~\ref{subsec:prelim}).
We then introduce \ours{}, describing the technical details on the dynamic encoder with target-time residual adaptation (\S~\ref{subsec:ours}).
Finally, we detail the training strategy (\S~\ref{subsec:training}) and conclude by analyzing the learned residual features (\S~\ref{sec:analysis}).

\input{figures/arch}

\subsection{Preliminaries: Latent-based Novel View Synthesis}
\label{subsec:prelim}

Latent-based static NVS methods~\citep{szymanowicz2026lagernvs, jiang2025rayzer} encode multi-view images of a static scene into a latent scene representation and synthesize a novel view by decoding this latent conditioned on a target camera pose. Such a framework comprises a multi-view encoder $\mathcal{E}_\text{st}$ and a renderer $\mathcal{R}$. Camera parameters are optional, conditioning the encoder when available and otherwise letting the model operate pose-free.

Given $N$ input images $\{\mathbf{I}_k\}_{k=1}^N$ and their optional corresponding cameras $\{\pi_k\}_{k=1}^N$, the static encoder produces the static scene latent $\mathbf{z}_\text{st}$:
\begin{equation}
\mathbf{z}_\text{st} = \mathcal{E}_\text{st}\big(\{\mathbf{x}_k,\ \pi_k\}_{k=1}^N\big) \in \mathbb{R}^{N\times P\times C},
\qquad \mathbf{x}_k = \mathcal{P}_\text{st}(\mathbf{I}_k),
\end{equation}
with $\mathcal{P}_\text{st}(\cdot)$ denotes the patchify layer, $P$ the number of image patches, and $C$ the feature dimension. The renderer decodes the latent into a novel-view image $\hat{\mathbf{I}} = \mathcal{R}(\mathbf{z}_\text{st},\pi_\text{tgt})$ at a query camera $\pi_\text{tgt}$.

\subsection{\ours{}: Feed-forward Latent Model for 4D Novel View Synthesis}
\label{subsec:ours}
\paragrapht{Dynamic encoder.}
To learn to disentangle dynamic motion from static scene structure in monocular videos, we freeze the pretrained static encoder $\mathcal{E}_\text{st}$ and renderer $\mathcal{R}$, and introduce a lightweight dynamic encoder $\mathcal{E}_\text{dyn}$.
The static encoder $\mathcal{E}_\text{st}$ already captures time-invariant content and camera-induced viewpoint changes from its static-scene training.
During our dynamic-scene training, the frozen static encoder and renderer allow the dynamic encoder $\mathcal{E}_\text{dyn}$ to learn temporal variations as residuals to the pretrained static representation.
Freezing the renderer $\mathcal{R}$ fixes latent-to-image mapping, forcing the residual feature to the pretrained latent space. This encourages them to complement the static representation rather than learn a new feature space from limited dynamic data.

To provide explicit temporal cues to the dynamic encoder $\mathcal{E}_\text{dyn}$, we introduce a learnable time token $\boldsymbol{\tau}_k \in \mathbb{R}^{C}$ for each context frame $\mathbf{I}_k$, which is concatenated with the patch tokens $\mathbf{x}_k = \mathcal{P}_\text{dyn}(\mathbf{I}_k)$ before being fed to the dynamic encoder $\mathcal{E}_\text{dyn}$.
Through this encoding, each time token encodes a timestamp-related information from each frame.

In the dynamic encoder, identifying dynamic regions requires separating object motion from apparent camera-induced motion.
We therefore introduce \textit{static anchoring}, which conditions the dynamic encoder on static feature through cross-attention. This allows it to factor out camera-induced changes captured in the static features and focus its representation on residual motion from dynamic objects. 
Each dynamic encoder block applies frame-wise self-attention $\mathrm{FSA}(\cdot)$,
global self-attention $\mathrm{GSA}(\cdot)$, and cross-attention $\mathrm{CA}(\cdot)$ sequentially,
\begin{equation}
    \mathbf{F}_\text{res}^{\ell} \leftarrow \mathbf{F}_\text{res}^{\ell} + \mathrm{FSA}^{\ell}(\mathbf{F}_\text{res}^{\ell}),
    \quad
    \mathbf{F}_\text{res}^{\ell} \leftarrow \mathbf{F}_\text{res}^{\ell} + \mathrm{GSA}^{\ell}(\mathbf{F}_\text{res}^{\ell}),
    \quad
    \mathbf{F}_\text{res}^{\ell+1} = \mathbf{F}_\text{res}^{\ell} + \mathrm{CA}^{\ell}\big(\mathbf{F}_\text{res}^{\ell},\,
    \mathbf{F}_\text{st}\big),
\end{equation}
where $\mathbf{F}_\text{res}^{\ell}$ denotes the residual features at the $\ell$-th dynamic encoder block,
$\mathbf{F}_\text{st}$ the features from the static encoder, and
$\mathrm{CA}(X, Y)$ computes queries from $X$ and keys and values from
$Y$.

\paragrapht{Target-time residual adaptation.}
The residual latent and encoded time tokens $[\,\mathbf{z}_\text{res};\hat{\boldsymbol{\tau}}_1,\dots,\hat{\boldsymbol{\tau}}_N\,] = \mathcal{E}_\text{dyn}(\{\textbf{I}_k\}_{k=1}^N)\in\mathbb{R}^{N\times (P+1)\times C}$ summarize temporal information across all $N$ context frames.
To extract a residual specific to a target timestamp, we condition the residual latent $\mathbf{z}_\text{res}$ with target-time token $\hat{\boldsymbol{\tau}}_\text{tgt}$.
For an observed timestamp $t_\text{tgt}=t_k$, we directly use the corresponding token $\hat{\boldsymbol{\tau}}_k$.
For an intermediate, novel timestamp $t_\text{tgt}\in[t_k,t_{k+1}]$, we linearly interpolate the two time tokens adjacent to the target timestamp $(t_{k},t_{k+1})$, weighted by their temporal distance to the target. 
\begin{equation}
    \hat{\boldsymbol{\tau}}_\text{tgt} = \frac{(t_{k+1}-t_\text{tgt})\cdot \hat{\boldsymbol{\tau}}_{k} + (t_\text{tgt}-t_k)\cdot \hat{\boldsymbol{\tau}}_{k+1}}{t_{k+1} - t_k},
    \qquad t_\text{tgt}\in[t_k,\,t_{k+1}].
\end{equation}
The interpolated token then provides the temporal condition for rendering at the novel timestamp.

Given $\hat{\boldsymbol{\tau}}_\text{tgt}$, a lightweight time module $\mathcal{T}$ modulates the aggregated residuals $\mathbf{z}_\text{res}$ into the target-time residual $\mathbf{z}_\text{res}^{\text{tgt}}$. 
The time module $\mathcal{T}$ consists of two transformer blocks with adaptive layer normalization (AdaLN) conditioned on $\hat{\boldsymbol{\tau}}_\text{tgt}$, followed by \textit{a residual gate} implemented as a zero-initialized linear projection layer.
This gate initializes the residual contribution to zero (\ie, preserving the pretrained static prior at the start of training) and progressively learns temporal corrections.
The target residual $\mathbf{z}_\text{res}^{\text{tgt}}$ is then added to the static latent $\mathbf{z}_\text{st}$ to obtain the target-time dynamic latent $\mathbf{z}_\text{dyn}^\text{tgt}$:
\begin{equation}
\mathbf{z}_\text{res}^{\text{tgt}} = \mathcal{T}(\mathbf{z}_\text{res};\ \hat{\boldsymbol{\tau}}_\text{tgt}),
\qquad
\mathbf{z}_\text{dyn}^\text{tgt} = \mathbf{z}_\text{st} + \mathbf{z}_\text{res}^{\text{tgt}}.
\end{equation}
The residual $\mathbf{z}_\text{res}$ is a time-invariant representation, processed only once for the $N$ context frames by the dynamic encoder $\mathcal{E}_\text{dyn}$.
For multiple target timestamps, we reuse $\mathbf{z}_\text{res}$ and apply the lightweight time module $\mathcal{T}$ with corresponding target-time tokens. The resulting target-specific latent is then fed to the frozen renderer $\mathcal{R}$ to render an image at the query camera $\pi_\text{tgt}$:
\begin{equation}
\hat{\mathbf{I}} = \mathcal{R}(\mathbf{z}_\text{dyn}^\text{tgt}, \pi_\text{tgt}).
\end{equation}

\subsection{Training}
\label{subsec:training}

The rendered image $\hat{\mathbf{I}}$ at the target view and time is supervised by an image reconstruction loss:
\begin{equation}
\mathcal{L}_{\text{image}} = \mathcal{L}_\mathrm{2} + \lambda_\text{perc}\mathcal{L}_{\text{perc}} + \lambda_{\text{perc}}^\text{dyn}\mathcal{L}_{\text{perc}}^{\text{dyn}} + \lambda_\text{grad}^\text{dyn}\mathcal{L}_{\text{grad}}^{\text{dyn}},
\end{equation}
where the L2 loss $\mathcal{L}_\text{2}$ and the perceptual loss $\mathcal{L}_\text{perc}$ are computed over the full image, with an extra perceptual loss $\mathcal{L}_{\text{perc}}^\text{dyn}$ and the gradient loss $\mathcal{L}_{\text{grad}}^\text{dyn}$ on dynamic region.
Since moving objects typically occupy a small fraction of the image, the full-image losses are dominated by static pixels and provide weak supervision for them.
The dynamic-region losses provide stronger supervision for dynamic objects and yield sharper rendering of moving objects.

To help disentangle dynamic object motion from camera-induced motion without dense geometric or motion supervision, we indirectly supervise the dynamic encoder to locate dynamic regions.
Specifically, we attach a DPT-based mask head~\citep{ranftl2021visiondpt} to the dynamic encoder to predict a binary dynamic-region mask $\hat M_k$, which is supervised with a pseudo label of dynamic mask $M_k$ using a binary cross-entropy loss $\text{BCE}(\cdot)$,
\begin{equation}
\mathcal{L}_{\text{mask}} = \text{BCE}(\hat M_k, M_k).
\end{equation}
The total training objective is
\begin{equation}
\mathcal{L} = \mathcal{L}_{\text{image}} + \lambda_\text{mask}\mathcal{L}_{\text{mask}}.
\end{equation}

Further details of training objective and pseudo-label generation are provided in Appendix~\ref{supp:impl}.

\subsection{Analysis on residual latent}
\label{sec:analysis}

We conduct analyses to investigate what information is captured in the residual latent $\mathbf{z}_\text{res}$. Additional analysis on static and residual latent is provided in Appendix~\ref{supp:feat_analysis}.

\input{figures/compare_static}
\input{figures/dynfeat}
\input{figures/gatemag}

\paragrapht{\textbf{Residual complements static rendering.}}
The frozen static model can answer \emph{where} to render, but not \emph{when}.
Having no target-time input, the static latent $\mathbf{z}_\text{st}$ compresses the video sequence into a single time-invariant prediction; moving content is thus discarded or frozen at observed frames with the input pose entangled, as in \cref{fig:comparestatic}.
Our residual latent $\mathbf{z}_\text{res}$ supplies this missing temporal axis.
Conditioned on the target time, it complements the static latent with time-dependent variation, placing the dynamic object at its target-time position while preserving the static background.

\paragrapht{\textbf{Residual feature encodes temporal dynamics.}}
\Cref{fig:dynfeat} compares the static latent $\mathbf{z}_\text{st}$ and residual latent $\mathbf{z}_\text{res}$ using PCA visualization. While the static latent exhibits camera-ray-like patterns, the residual latent suppresses static regions and captures distinct motion across time, suggesting that the dynamic encoder effectively isolates temporal variation from the time-invariant scene representation.

This temporal specialization is further supported by the learned residual gate.
\Cref{fig:gate} visualizes gate activation using the channel-wise L2 norm of the gated residual. The gate remains largely inactive in static regions but activates strongly around moving objects, indicating that temporal modeling capacity is concentrated where motion occurs. 

%% file: figures/arch.tex
\begin{figure}[t]
\centering

\includegraphics[width=1.00\textwidth]{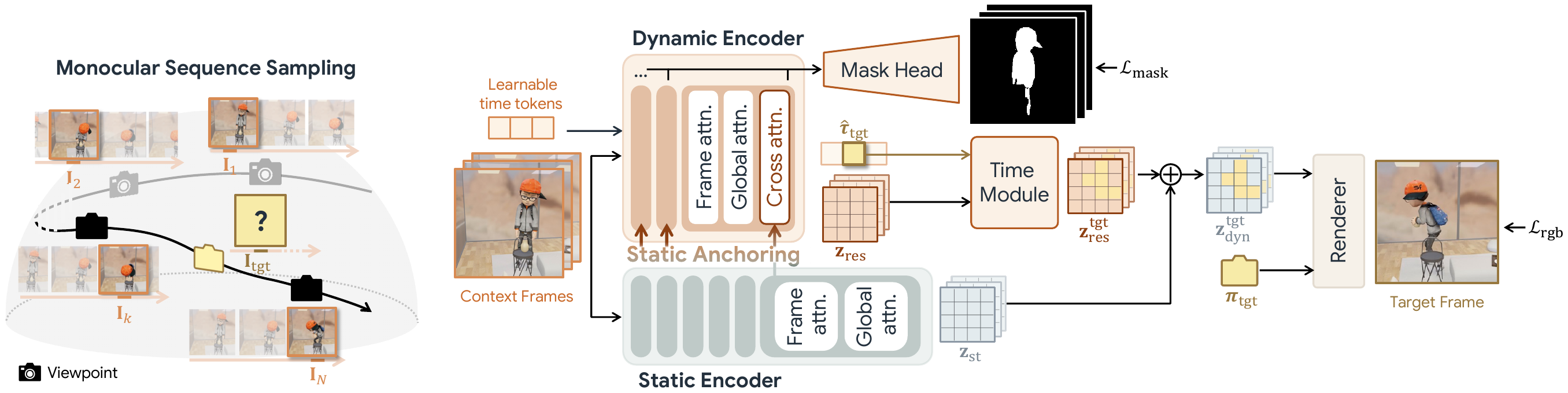}

\caption{\textbf{\ours{} Overview.}
\figleft{} During training, context and target frames with varying views and timestamps are sampled from multi-view videos. 
\figright{} A static encoder encodes static 3D structures (static latent) while a lightweight dynamic encoder extracts temporal residuals (residual latent) within the whole context frames.
For 4D NVS, the residual latent is first adapted at the target time and added to the static latent resulting in a dynamic latent, which is then rendered into the novel view with target-view conditioned.
}
\label{fig:arch}
\afterfig
\end{figure}

%% file: figures/compare_static.tex
\begin{figure}[t]
\centering

\includegraphics[width=1.00\textwidth]{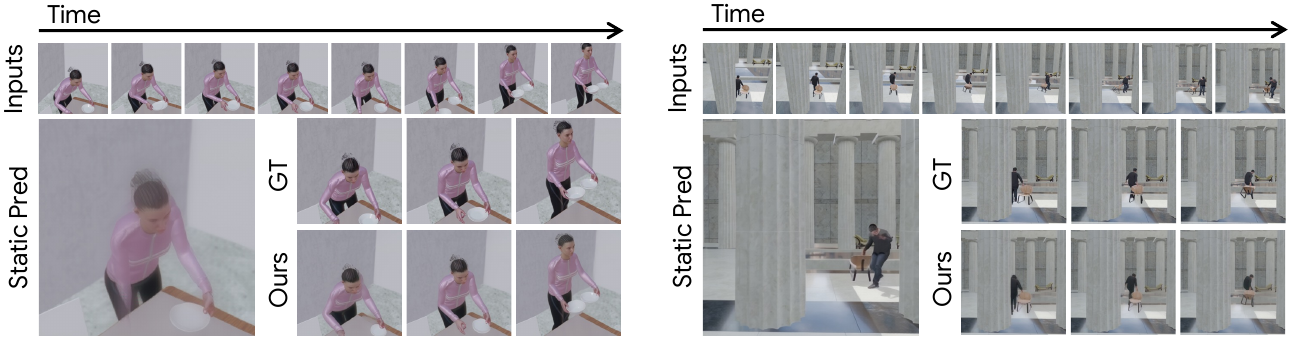}

\beforetfigcaption

\caption{\textbf{Rendered frames from the static latent and our residual time correction.} In dynamic scenes, the static latent suppresses dynamic content (left) or copies it from the nearest context frame (right), whereas our residual latent places the moving content at the target time while preserving the static background.}
\label{fig:comparestatic}
\afterfig
\end{figure}

%% file: figures/dynfeat.tex
\begin{figure}[t]
\centering

\includegraphics[width=1.00\textwidth]{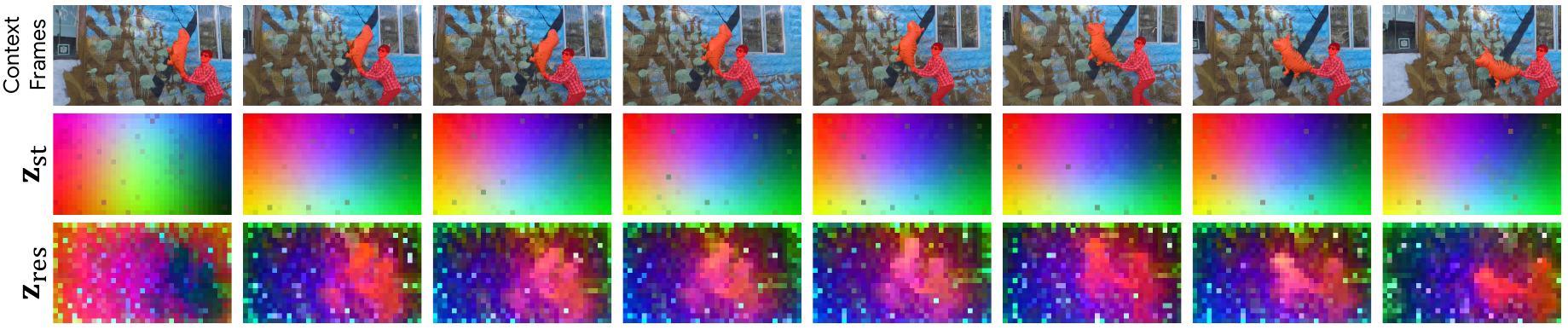}

\beforetfigcaption

\caption{\textbf{PCA visualization of static and residual latent.} The static latent captures camera-ray-like structure, whereas the residual latent primarily attends to moving objects and exhibits temporal variation.}
\label{fig:dynfeat}
\afterfig
\end{figure}

%% file: figures/gatemag.tex
\begin{figure}[!tp]
\centering

\includegraphics[width=1.00\textwidth]{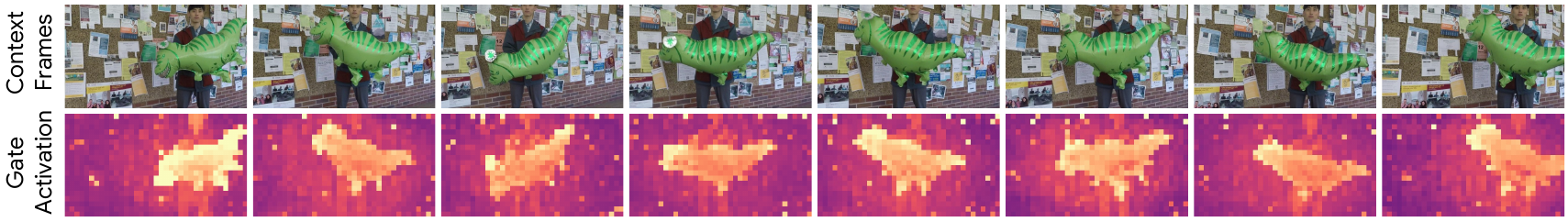}

\beforetfigcaption

\caption{\textbf{Visualization of residual gate activation.} The residual gate magnitude is higher on dynamic objects while remaining suppressed in the background.}
\label{fig:gate}
\afterfig
\end{figure}

%% file: sections/4_experiments.tex
\section{Experiments}
\label{sec:exp}

\paragrapht{Implementation details.}
Our model builds on a frozen static novel view synthesizer~\citep{szymanowicz2026lagernvs}, with only the dynamic encoder, time module, and mask head trained. The dynamic encoder follows the architecture of the pretrained static encoder, with added cross-attention module for each layer but uses one-third of its depth, with 8 transformer layers compared to 24 in the static encoder, since temporal variation accounts for only a relatively small portion of the scene. Further architectural and training details are provided in Appendix~\ref{supp:impl}.

\paragrapht{\textbf{Training dataset.}} 
To effectively disentangle camera and object motion, we train our model using synchronized multi-view video datasets such as Kubric4D~\citep{van2024generativekub4d} and CamxTime~\citep{huang2025spacetimepilotcamxtime}, captured by $V$ cameras with fixed poses with $T$ frames.
Unlike costly, explicit 3D annotation (\eg, dense depth or 3D motion) that is commonly used for 3D-Gaussian-based approaches, the multi-view datasets provide cost-effective indirect 4D supervision from 2D images. 
Furthermore, our residual formulation specifically makes this RGB-only supervision directly tractable, with requiring relatively small amount as well.

Given a multi-view video $\mathcal{I}= \{I_{v,t}\mid v\in\{1,\cdots,V\}, t\in\{1,\cdots,T\}\}$, we adopt an asynchronous sampling strategy to ensure generalization to monocular video inputs at test time as well as accurate conditioning on arbitrary viewpoints and timestamps.
We sample $N$ distinct view-time pairs $\{(v_i,t_i)\}_{i=1}^{N}\sim\mathcal{U}\left(\{1,\cdots,V\}\times\{1,\cdots,T\}\right)$, then construct an asynchronous input set $\mathcal{I}_\text{in}=\left\{I_{v_i,t_i}\right\}_{i=1}^{N}$. We then sample the target frames from the remaining set such that $\mathcal{I}_\text{tar} \subseteq \mathcal{I} \setminus \mathcal{I}_\text{in}$.
Holding the viewpoint fixed isolates temporal changes caused by dynamic objects, whereas fixing the timestamp provides observations from varying viewpoints, enabling supervision analogous to static multi-view synthesis. This setup naturally trains ours to disentangle temporal dynamics from viewpoint changes.

\input{figures/eval_settings}
\input{tables/NVS}
\input{tables/CTVS}

\paragrapht{Evaluation setting.}
We evaluate three settings, novel view synthesis (NVS), cross-time view synthesis (CTVS), and novel time synthesis (NTS), illustrated in~\cref{fig:evalsettings}.
NVS evaluates held-out viewpoints at input timestamps, CTVS evaluates unseen viewpoints-timestamps combinations from a monocular video, and NTS evaluates intermediate timestamps from temporally strided inputs.
NVS and CTVS use multi-view datasets from fixed cameras: NVIDIA~\citep{yoon2020novelnvidiadataset}, N3DV~\citep{li2022neuraln3dv}, Kubric4D, CamxTime, and Charge~\citep{nazarczuk2026charge}. 
NTS is evaluated on monocular videos captured by moving cameras: DyCheck~\citep{gao2022monoculardycheck}, ADT~\citep{pan2023ariaadt}, and TUM~\citep{sturm2012benchmarktum}.
Kubric4D and CamxTime are evaluated on held-out test scenes from our training distribution, whereas the other datasets evaluate zero-shot generalization. Detailed evaluation protocol is provided in Appendix~\ref{supp:evalproto}.

\input{tables/NTS}

\subsection{Results}
\label{sec:results}

\paragrapht{Quantitative results.}
Tables~\ref{tab:nvs},~\ref{tab:ctvs}, and~\ref{tab:nts} report the three protocols. Here, $^\dagger$ denotes 2{,}000-step test-time pose alignment with target frames, and Pose (\cmark/\xmark) indicates context pose input. 
\ours{} achieves state-of-the-art performance among all pose-free baselines, even without test-time alignment, except for SSIM on NVS of two real-world datasets.
On real-world datasets, whose camera poses are noisy estimates, C4G benefits from test-time pose alignment while our model is forced to estimate slight inaccurate pose relation, resulting in a slight misalignment of pixels.
An optional incorporation of context camera poses further improves the accuracy by exploiting the explicit context-to-target pose relation achieving state-of-the-art performance across all benchmarks on all protocols.
These results demonstrate the substantial strength of our proposed latent representation: strong 4D prior, generalization to novel view and time, and decouplement of view and time, all successfully learned in the latent space.
Additional rendering quality evaluation on dynamic region and discussion on test-time pose alignment are provided in Appendices~\ref{supp:dynregion} and ~\ref{supp:tto}.

\input{tables/ablation}
\paragrapht{Ablation.} \label{sec:ablation}
\Cref{tab:ablation} ablates our key design choices to justify our formulation, with all variants trained for 10k steps. 
We also include metrics on dynamic region to see its gain exclusively.
The residual branch denotes our newly introduced modules (\ie, the dynamic encoder and the time module) on top of the static NVS model.
Additional ablation studies are provided in Appendix~\ref{supp:arch_ablation}.

Variant (i) is the static NVS model baseline, showing huge regression from \ours{}.
Freezing the renderer is crucial (variant (ii)); unfreezing it disrupts dynamic branch training by destabilizing the established latent-to-image mapping in the renderer.
The dynamic mask loss (variant (iii)) modestly improves the accuracy by helping the model resolve depth ambiguity and localize dynamic regions during training. Without it, the model tends to collapse as training progresses.
Static feature anchoring to the residual encoder (variant (iv)) further improves the overall accuracy, helping the residual branch to effectively isolate temporal variation from injected intermediate features from the static encoder.

\input{tables/runtime}
\paragrapht{Runtime.}
\Cref{tab:runtime} reports runtime on the NVIDIA dataset at $224\times224$ resolution using a single RTX 5880 Ada GPU.
\ours{} achieves the fastest runtime.
Ours is 1.08$\times$ faster than NeoVerse and 565$\times$ faster than that with pose alignment, with ours achieving substantially better accuracy.

\input{figures/NVS_qual}
\paragrapht{Qualitative results.} \label{sec:qual}
\Cref{fig:nvsqual} qualitatively compares our results with Gaussian-based methods.
Gaussian-based methods often exhibit holes, object inconsistencies, and blurred details when rendering at novel viewpoints and timestamps. In contrast, our method produces visually consistent results.
This again confirms the superiority of our latent-based method over explicit Gaussian primitives: 4D prior and view-time disentanglement can be learned successfully in the latent space, and novel view are seamlessly rendered.
Appendix~\ref{supp:qual} includes comprehensive qualitative comparisons on all benchmark datasets. 

\subsection{Latent probing on downstream tasks}
\label{sec:probing}
To assess whether the dynamic encoder captures temporal changes while preserving scene geometry encoded by static feature, we probe the representations via zero-shot temporal matching and video-depth probing. In tables, ``All'' covers the entire frame, while ``Dyn'' is restricted to dynamic objects. Full matching and depth results, along with point-tracking results, are provided in Appendix~\ref{supp:probing}.

\input{tables/probe_matching_main}

\paragrapht{Zero-shot matching.}
While the static prior establishes dense correspondences across views, exploiting epipolar geometry, moving content violates the multi-view consistency, degrading correspondence quality in dynamic regions.
On the other hand, dynamic prior improves correspondence in such regions.
In \cref{tab:probe_matching}, we evaluate zero-shot matching of static ($\mathbf{z}_\text{st}$) and dynamic latent ($\mathbf{z}_\text{dyn}$) on DAVIS, using the protocol of \citet{nam2026emergentdifftrack}.
The dynamic latent consistently demonstrates better accuracy than the static latent; the gap slightly widens on dynamic region.

\input{tables/probe_depth_main}

\paragrapht{Video depth estimation.}
Although the model receives no explicit geometric supervision, its latent representations can encode scene geometry information. We probe this by attaching DPT depth predictors on the static and dynamic latent and training them Kubric~\citep{greff2022kubric}.
\Cref{tab:probe_depth} reports the video depth accuracy. Dynamic latent consistently outperforms static latent, across the full frame and dynamic-object regions. Notably, the advantage becomes more pronounced as the temporal stride between context frames increases, corresponding to larger object motion. This suggests that dynamic latent largely preserves geometry-aware information encoded by the static latent while incorporating additional cues from temporal dynamics.

%% file: figures/eval_settings.tex
\begin{figure}[t]
\centering

\includegraphics[width=1.0\linewidth]{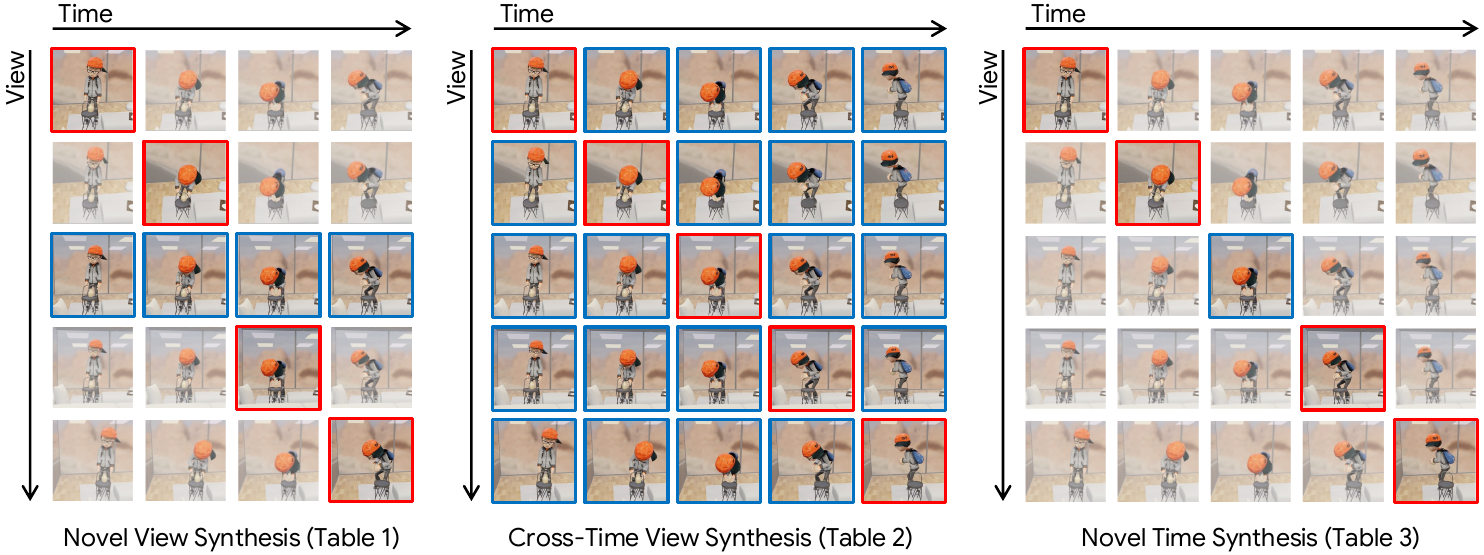}

\caption{\textbf{Three evaluation settings} given synchronized multi-view video datasets. \textcolor[HTML]{FF0000}{Red} outlines indicate input context frames, while \textcolor[HTML]{0070C0}{blue} outlines indicate target frames for evaluation.}
\label{fig:evalsettings}
\afterfig
\end{figure}

%% file: tables/NVS.tex
\begin{table}[t]
\centering
\caption{\textbf{Quantitative results on Novel View Synthesis:}
\ours{} outperforms all Gaussian-based methods on all benchmark datasets, showing the strength of our latent representation bypassing explicit primitives.
Context pose input further improves the metrics, achieving the state of the art.
\firstbox{Best}, \secondbox{second best}, and \thirdbox{third best} results are highlighted.}
\aftertabcaption
\label{tab:nvs}
\resizebox{\textwidth}{!}{%
\begin{tabular}{@{}l c ccc ccc ccc ccc ccc@{}}
\toprule
\multirow{2}{*}{Method} & \multirow{2}{*}{Pose}
& \multicolumn{3}{c}{NVIDIA} & \multicolumn{3}{c}{N3DV} & \multicolumn{3}{c}{Kubric4D} & \multicolumn{3}{c}{CamxTime} & \multicolumn{3}{c}{Charge} \\
\cmidrule(lr){3-5}\cmidrule(lr){6-8}\cmidrule(lr){9-11}\cmidrule(lr){12-14}\cmidrule(lr){15-17}
& & PSNR$\uparrow$ & SSIM$\uparrow$ & LPIPS$\downarrow$ & PSNR$\uparrow$ & SSIM$\uparrow$ & LPIPS$\downarrow$ & PSNR$\uparrow$ & SSIM$\uparrow$ & LPIPS$\downarrow$ & PSNR$\uparrow$ & SSIM$\uparrow$ & LPIPS$\downarrow$ & PSNR$\uparrow$ & SSIM$\uparrow$ & LPIPS$\downarrow$ \\
\midrule
\multicolumn{8}{@{}l}{\emph{Feed-forward Gaussian Representation}}\\

4DGT~\citep{4dgt} & \cmark & 16.127 & 0.3884 & 0.5972 & 17.813 & 0.6609 & 0.5131 & 14.148 & 0.5027 & 0.6096 & 17.125 & 0.6557 & 0.4893 & 18.017 & 0.7786 & 0.5653 \\
MoVieS~\citep{lin2026movies} & \cmark & 11.818 & 0.2546 & 0.6760 & \cellcolor{tabsecond}20.136 & \cellcolor{tabthird}0.7045 & \cellcolor{tabsecond}0.4605 & 12.411 & 0.3550 & \cellcolor{tabthird}0.6070 & 17.069 & 0.6223 & 0.4488 & \cellcolor{tabthird}18.709 & 0.7473 & 0.5298 \\
NeoVerse$^\dagger$~\citep{yang2026neoverse} & \xmark & 15.050 & \cellcolor{tabthird}0.4135 & \cellcolor{tabsecond}0.4626 & 12.953 & 0.4806 & 0.5697 & 8.954 & 0.2573 & 0.6721 & 10.432 & 0.3110 & 0.5905 & 15.224 & 0.5211 & 0.5919 \\
C4G$^\dagger$~\citep{c4g} & \xmark & \cellcolor{tabthird}16.572 & \cellcolor{tabsecond}0.4182 & 0.6203 & 18.444 & \cellcolor{tabsecond}0.7050 & 0.5246 & \cellcolor{tabthird}14.931 & \cellcolor{tabthird}0.5656 & 0.6442 & \cellcolor{tabthird}20.527 & \cellcolor{tabthird}0.7056 & \cellcolor{tabthird}0.4369 & 18.535 & \cellcolor{tabthird}0.7797 & \cellcolor{tabthird}0.5279 \\
\midrule
\multicolumn{8}{@{}l}{\emph{Feed-forward Latent-based 4D Representation}}\\
\textbf{\ours{} (Ours)} & \xmark & \cellcolor{tabsecond}17.945 & 0.4113 & \cellcolor{tabthird}0.5229 & \cellcolor{tabthird}18.581 & 0.6898 & \cellcolor{tabthird}0.4776 & \cellcolor{tabsecond}20.916 & \cellcolor{tabsecond}0.6433 & \cellcolor{tabsecond}0.3767 & \cellcolor{tabsecond}21.268 & \cellcolor{tabsecond}0.7237 & \cellcolor{tabsecond}0.2743 & \cellcolor{tabsecond}20.924 & \cellcolor{tabsecond}0.8217 & \cellcolor{tabsecond}0.4952 \\
\textbf{\ours{} (Ours)} & \cmark & \cellcolor{tabfirst}20.435 & \cellcolor{tabfirst}0.4835 & \cellcolor{tabfirst}0.3867 & \cellcolor{tabfirst}22.896 & \cellcolor{tabfirst}0.7716 & \cellcolor{tabfirst}0.3560 & \cellcolor{tabfirst}22.209 & \cellcolor{tabfirst}0.6737 & \cellcolor{tabfirst}0.3350 & \cellcolor{tabfirst}25.730 & \cellcolor{tabfirst}0.8270 & \cellcolor{tabfirst}0.1681 & \cellcolor{tabfirst}22.959 & \cellcolor{tabfirst}0.8324 & \cellcolor{tabfirst}0.4616 \\

\bottomrule
\end{tabular}%
}
\aftertab
\end{table}

%% file: tables/CTVS.tex
\begin{table}[!t]
\centering
\caption{On the challenging \textbf{Cross-Time View Synthesis} task, 
\ours{} substantially outperforms all Gaussian-based methods on all benchmark datasets, even without test-time pose alignment ($^\dagger$), showing the model successfully learns 4D priors and decouples view and time in the latent space.
\firstbox{Best}, \secondbox{second best}, and \thirdbox{third best} results are highlighted.}
\aftertabcaption
\label{tab:ctvs}
\resizebox{\textwidth}{!}{%
\begin{tabular}{@{}l c ccc ccc ccc ccc ccc@{}}
\toprule
\multirow{2}{*}{Method} & \multirow{2}{*}{Pose}
& \multicolumn{3}{c}{NVIDIA} & \multicolumn{3}{c}{N3DV} & \multicolumn{3}{c}{Kubric4D} & \multicolumn{3}{c}{CamxTime} & \multicolumn{3}{c}{Charge} \\
\cmidrule(lr){3-5}\cmidrule(lr){6-8}\cmidrule(lr){9-11}\cmidrule(lr){12-14}\cmidrule(lr){15-17}
& & PSNR$\uparrow$ & SSIM$\uparrow$ & LPIPS$\downarrow$ & PSNR$\uparrow$ & SSIM$\uparrow$ & LPIPS$\downarrow$ & PSNR$\uparrow$ & SSIM$\uparrow$ & LPIPS$\downarrow$ & PSNR$\uparrow$ & SSIM$\uparrow$ & LPIPS$\downarrow$ & PSNR$\uparrow$ & SSIM$\uparrow$ & LPIPS$\downarrow$ \\
\midrule
\multicolumn{8}{@{}l}{\emph{Feed-forward Gaussian Representation}}\\

4DGT~\citep{4dgt} & \cmark & 15.937 & 0.3867 & 0.6052 & 16.097 & 0.6380 & 0.5483 & 15.680 & 0.5424 & 0.5786 & 17.024 & 0.6573 & 0.4939 & 17.939 & 0.7756 & 0.5617 \\
MoVieS~\citep{lin2026movies} & \cmark & 15.785 & \cellcolor{tabthird}0.4614 & 0.5328 & \cellcolor{tabthird}19.710 & \cellcolor{tabthird}0.7400 & \cellcolor{tabthird}0.4414 & \cellcolor{tabthird}16.988 & \cellcolor{tabthird}0.6368 & \cellcolor{tabthird}0.4174 & 17.340 & 0.6470 & \cellcolor{tabthird}0.4224 & \cellcolor{tabthird}20.113 & \cellcolor{tabthird}0.7771 & \cellcolor{tabthird}0.5084 \\
NeoVerse$^\dagger$~\citep{yang2026neoverse} & \xmark & 14.449 & 0.3819 & \cellcolor{tabthird}0.4890 & 12.447 & 0.4729 & 0.5681 & 8.885 & 0.2256 & 0.6793 & 10.276 & 0.3091 & 0.5911 & 15.220 & 0.5380 & 0.5940 \\
C4G$^\dagger$~\citep{c4g} & \xmark & \cellcolor{tabthird}16.403 & 0.4096 & 0.6304 & 17.039 & 0.6829 & 0.5374 & 15.270 & 0.5572 & 0.6348 & \cellcolor{tabthird}20.175 & \cellcolor{tabthird}0.7009 & 0.4453 & 18.410 & 0.7640 & 0.5324 \\
\midrule
\multicolumn{8}{@{}l}{\emph{Feed-forward Latent-based 4D Representation}}\\
\textbf{\ours{} (Ours)} & \xmark & \cellcolor{tabsecond}21.552 & \cellcolor{tabsecond}0.5348 & \cellcolor{tabsecond}0.3739 & \cellcolor{tabsecond}24.550 & \cellcolor{tabsecond}0.8152 & \cellcolor{tabsecond}0.3353 & \cellcolor{tabsecond}23.547 & \cellcolor{tabsecond}0.7535 & \cellcolor{tabsecond}0.2558 & \cellcolor{tabsecond}24.546 & \cellcolor{tabsecond}0.8040 & \cellcolor{tabsecond}0.2016 & \cellcolor{tabsecond}24.222 & \cellcolor{tabfirst}0.8500 & \cellcolor{tabsecond}0.4413 \\
\textbf{\ours{} (Ours)} & \cmark & \cellcolor{tabfirst}21.838 & \cellcolor{tabfirst}0.5578 & \cellcolor{tabfirst}0.3562 & \cellcolor{tabfirst}25.109 & \cellcolor{tabfirst}0.8284 & \cellcolor{tabfirst}0.3234 & \cellcolor{tabfirst}24.272 & \cellcolor{tabfirst}0.7850 & \cellcolor{tabfirst}0.2304 & \cellcolor{tabfirst}27.648 & \cellcolor{tabfirst}0.8732 & \cellcolor{tabfirst}0.1325 & \cellcolor{tabfirst}24.421 & \cellcolor{tabsecond}0.8496 & \cellcolor{tabfirst}0.4382 \\

\bottomrule
\end{tabular}%
}
\aftertab
\end{table}

%% file: tables/NTS.tex
\begin{table}[!t]
\centering
\tiny
\setlength\tabcolsep{4.2pt}
\caption{On \textbf{Novel Time Synthesis},
\ours{} achieves the state-of-the-art accuracy by a large margin over all methods, showing the strong generalization of our latent-based approach at novel interpolated timestamps.
\firstbox{Best}, \secondbox{second best}, and \thirdbox{third best} results are highlighted.}
\aftertabcaption
\label{tab:nts}
\begin{tabularx}{0.9\columnwidth}{@{} X c ccc ccc ccc @{}}
\toprule
\multirow{2}{*}{Method} & \multirow{2}{*}{Pose}
& \multicolumn{3}{c}{DyCheck} & \multicolumn{3}{c}{ADT} & \multicolumn{3}{c}{TUM} \\
\cmidrule(lr){3-5}\cmidrule(lr){6-8}\cmidrule(lr){9-11}
& & PSNR$\uparrow$ & SSIM$\uparrow$ & LPIPS$\downarrow$ & PSNR$\uparrow$ & SSIM$\uparrow$ & LPIPS$\downarrow$ & PSNR$\uparrow$ & SSIM$\uparrow$ & LPIPS$\downarrow$ \\
\midrule
\multicolumn{8}{@{}l}{\emph{Feed-forward Gaussian Representation}}\\

4DGT~\citep{4dgt} & \cmark & 13.743 & 0.4079 & 0.6233 & 16.710 & 0.4999 & 0.5651 & 14.380 & 0.4468 & 0.5612 \\
MoVieS~\citep{lin2026movies} & \cmark & 11.755 & 0.3420 & 0.6428 & 15.432 & 0.4698 & 0.5828 & 15.683 & 0.5079 & 0.4426 \\
NeoVerse$^\dagger$~\citep{yang2026neoverse} & \xmark & \cellcolor{tabthird}16.756 & 0.4663 & \cellcolor{tabthird}0.4566 & 16.179 & 0.4384 & \cellcolor{tabthird}0.5296 & 16.120 & 0.5266 & \cellcolor{tabthird}0.3981 \\
C4G$^\dagger$~\citep{c4g} & \xmark & 15.755 & \cellcolor{tabthird}0.4735 & 0.6111 & \cellcolor{tabthird}18.900 & \cellcolor{tabthird}0.5813 & 0.5833 & \cellcolor{tabthird}17.508 & \cellcolor{tabthird}0.5518 & 0.4951 \\
\midrule
\multicolumn{8}{@{}l}{\emph{Feed-forward Latent-based 4D Representation}}\\
\textbf{\ours{} (Ours)} & \xmark & \cellcolor{tabsecond}19.019 & \cellcolor{tabsecond}0.4977 & \cellcolor{tabsecond}0.4284 & \cellcolor{tabsecond}19.693 & \cellcolor{tabsecond}0.5948 & \cellcolor{tabsecond}0.4490 & \cellcolor{tabsecond}17.847 & \cellcolor{tabsecond}0.5586 & \cellcolor{tabsecond}0.3449 \\
\textbf{\ours{} (Ours)} & \cmark & \cellcolor{tabfirst}20.527 & \cellcolor{tabfirst}0.5470 & \cellcolor{tabfirst}0.3757 & \cellcolor{tabfirst}24.115 & \cellcolor{tabfirst}0.6875 & \cellcolor{tabfirst}0.3083 & \cellcolor{tabfirst}21.162 & \cellcolor{tabfirst}0.6978 & \cellcolor{tabfirst}0.2601 \\

\bottomrule
\end{tabularx}%
\aftertab
\vspace{-2em}
\end{table}

%% file: tables/ablation.tex
\begin{wraptable}[11]{r}{0.604\textwidth}
\centering
\vspace{-12pt}
\caption{\textbf{Ablation studies} on Kubric4D, on full frames and dynamic region. Each component contributes to our final formulation.}
\tiny
\setlength\tabcolsep{1pt}
\vspace{-8pt}
\label{tab:ablation}
\begin{tabularx}{0.604\columnwidth}{
@{} X cccc 
S[table-format=2.2,round-precision=2] S[table-format=1.3] S[table-format=1.3]
S[table-format=2.2, round-precision=2] S[table-format=1.3] S[table-format=1.3]
@{}}
\toprule
\multirow{4}{*}{Variant} & \multicolumn{4}{c}{Components} & \multicolumn{3}{c}{Full frame} & \multicolumn{3}{c}{Dynamic region} \\
\cmidrule(lr){2-5}\cmidrule(lr){6-8}\cmidrule(lr){9-11}
& {\makecell{Residual\\branch}} & {\makecell{Static\\anchoring}} & {\makecell{Mask\\loss}} & {\makecell{Frozen\\renderer}}
& PSNR$\uparrow$ & SSIM$\uparrow$ & LPIPS$\downarrow$ & PSNR$\uparrow$ & SSIM$\uparrow$ & LPIPS$\downarrow$ \\
\midrule

(i) & \xmark & -- & \xmark & \xmark & 21.902 & 0.7272 & 0.2872 & 13.161 & 0.4851 & 0.4375 \\
\midrule
(ii) & \cmark & \cmark & \cmark & \xmark & 22.147 & 0.6807 & 0.3095 & 13.828 & 0.4855 & 0.4647 \\
(iii) & \cmark & \cmark & \xmark & \cmark & 22.929 & 0.7405 & 0.2623 & 14.112 & 0.5002 & 0.4288 \\
(iv) & \cmark & \xmark & \cmark & \cmark & 23.608 & 0.7692 & 0.2461 & 14.253 & 0.5169 & 0.4363 \\
\midrule
\textbf{\ours{}} & \cmark & \cmark & \cmark & \cmark & \bfseries 23.666 & \bfseries 0.7734 & \bfseries 0.2369 & \bfseries 14.264 & \bfseries 0.5192 & \bfseries 0.4189 \\
\bottomrule
\end{tabularx}%
\end{wraptable}

%% file: tables/runtime.tex
\begin{wraptable}[8]{r}{0.32\textwidth}
\centering
\vspace{-22pt}
\tiny
\setlength\tabcolsep{2pt}
\caption{\textbf{Runtime.} $^\dagger$ denotes 2{,}000 step test-time pose alignment.}
\label{tab:runtime}
\vspace{-5pt}
\begin{tabularx}{0.32\textwidth}{@{}X ccc@{}}
\toprule
Method & sec / view $\downarrow$ & FPS $\uparrow$ & PSNR $\uparrow$ \\
\midrule
4DGT & 0.040 & 25.00 & 12.15 \\
MoVieS & 0.037 & 27.02 & 14.10 \\
NeoVerse & 0.027 & 37.03 & 10.74 \\
NeoVerse$^\dagger$ & 14.127 & 0.07 & 13.86 \\
C4G & 0.036 & 27.77 & 14.25 \\
C4G$^\dagger$ & 23.885 & 0.04 & 16.76 \\
\midrule
\textbf{\ours{} (Ours)} & \textbf{0.025} & \textbf{40.00} & \textbf{17.65} \\
\bottomrule
\end{tabularx}%
\end{wraptable}

%% file: figures/NVS_qual.tex
\begin{figure}[t]
\centering

\includegraphics[width=1.0\textwidth]{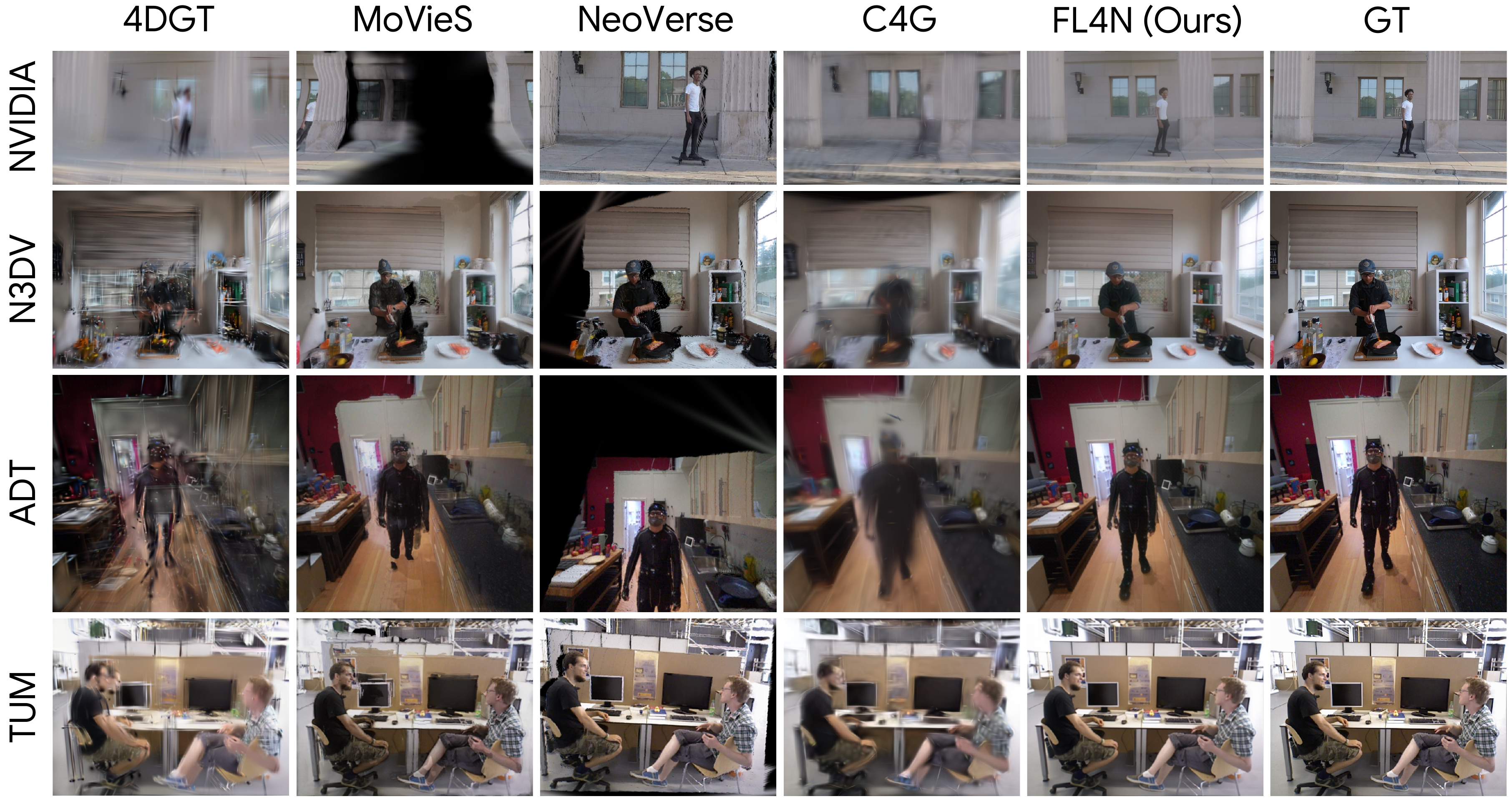}
\vspace{-5pt} 

\caption{\textbf{Qualitative comparison on 4D Novel View Synthesis.}
Gaussian-based methods often exhibit holes, object inconsistencies, and blurred details. Our method produces more coherent and visually consistent results, demonstrating the superiority of our latent-based representation over explicit geometry primitives.
}
\label{fig:nvsqual}
\afterfig
\vspace{-8pt}
\end{figure}

%% file: tables/probe_matching_main.tex
\begin{wraptable}[10]{r}{0.38\textwidth}
\centering

\vspace{-12.5pt}

\scriptsize
\setlength\tabcolsep{3pt}
\caption{
\textbf{Zero-shot matching on DAVIS.}
Dynamic latent outperform static latent by encoding temporal information.
}
\vspace{-6pt}
\label{tab:probe_matching}
\begin{tabularx}{0.38\textwidth}{@{}ll c c c c@{}}
\toprule
Region & Feat. & $\delta_\text{avg}$ & $\delta$1 & $\delta$4 & $\delta$16 \\
\midrule
\multirow{2}{*}{All}
& $\mathbf{z}_\text{st}$ & 27.33 & 5.65 & 25.37 & 53.40 \\
& $\mathbf{z}_\text{dyn}$ & \textbf{29.04} & \textbf{5.76} & \textbf{26.50} & \textbf{57.64} \\
&& {\tiny\textcolor{mygreen}{($+$1.71)}}
 & {\tiny\textcolor{mygreen}{($+$0.11)}}
 & {\tiny\textcolor{mygreen}{($+$1.13)}}
 & {\tiny\textcolor{mygreen}{($+$4.24)}} \\
\midrule
\multirow{2}{*}{Dyn}
& $\mathbf{z}_\text{st}$ & 23.45 & 3.10 & 21.14 & 49.03 \\
& $\mathbf{z}_\text{dyn}$ & \textbf{25.29} & \textbf{3.12} & \textbf{22.37} & \textbf{53.68} \\
&& {\tiny\textcolor{mygreen}{($+$1.84)}}
 & {\tiny\textcolor{mygreen}{($+$0.02)}}
 & {\tiny\textcolor{mygreen}{($+$1.23)}}
 & {\tiny\textcolor{mygreen}{($+$4.65)}} \\
\bottomrule
\end{tabularx}%

\end{wraptable}

%% file: tables/probe_depth_main.tex
\begin{wraptable}[10]{r}{0.4\textwidth}
\centering
\vspace{-11pt}

\caption{\textbf{Probing video depth on Kubric.} Dynamic latent yields consistent gains, with larger gains at greater temporal strides.}
\label{tab:probe_depth}
\vspace{-3pt}
\resizebox{\linewidth}{!}{%
\begin{tabular}{@{}ll cc cc@{}}
\toprule
\multirow{2}{*}{Region} & \multirow{2}{*}{Feat.}
& \multicolumn{2}{c}{$\Delta$4} & \multicolumn{2}{c}{$\Delta$8} \\
\cmidrule(lr){3-4}\cmidrule(lr){5-6}
&& AbsRel$\downarrow$ & $\delta<1.25\uparrow$ & AbsRel$\downarrow$ & $\delta<1.25\uparrow$ \\
\midrule
\multirow{3}{*}{All}
& $\mathbf{z}_\text{st}$ & 0.0944 & 0.8908 & 0.0868 & 0.9081 \\
& $\mathbf{z}_\text{dyn}$ & \textbf{0.0863} & \textbf{0.9035} & \textbf{0.0751} & \textbf{0.9326} \\
&& {\scriptsize\textcolor{mygreen}{($-$0.0081)}}
 & {\scriptsize\textcolor{mygreen}{($+$0.0127)}}
 & {\scriptsize\textcolor{mygreen}{($-$0.0117)}}
 & {\scriptsize\textcolor{mygreen}{($+$0.0244)}} \\
\midrule
\multirow{3}{*}{Dyn}
& $\mathbf{z}_\text{st}$ & 0.1178 & 0.8699 & 0.1149 & 0.8790 \\
& $\mathbf{z}_\text{dyn}$ & \textbf{0.1050} & \textbf{0.8941} & \textbf{0.0996} & \textbf{0.9099} \\
&& {\scriptsize\textcolor{mygreen}{($-$0.0128)}}
 & {\scriptsize\textcolor{mygreen}{($+$0.0242)}}
 & {\scriptsize\textcolor{mygreen}{($-$0.0153)}}
 & {\scriptsize\textcolor{mygreen}{($+$0.0309)}} \\
\bottomrule
\end{tabular}%
}
\end{wraptable}

%% file: sections/5_conclusion.tex
\section{Conclusion}
\label{sec:conclusion}

In this work, we introduced \ours{}, a feed-forward latent model that renders dynamic scenes from a monocular video at novel views and timestamps.
By learning a temporal residual on top of a frozen static latent model, \ours{} acquires a 4D representation from limited dynamic data, supervised only by images and binary dynamic masks.
Across eight benchmarks and three evaluation protocols, \ours{} outperforms prior feed-forward Gaussian-based methods and generalizes to real-world videos, while its latent space successfully disentangles view from time and preserves scene geometry.
Our work demonstrates the potential of latent representations for feed-forward 4D novel view synthesis, opening new directions for latent-based 4D scene understanding and modeling more complex dynamics.

%% file: sections/Appendix.tex
\appendix
\setcounter{table}{0}
\setcounter{figure}{0}
\renewcommand{\thetable}{\Alph{table}}
\renewcommand{\thefigure}{\Alph{figure}}

\etocsettocdepth.toc{subsection}
\etocsettocstyle{\section*{Appendix contents}}{}
\tableofcontents

\clearpage

\section{Implementation details}
\label{supp:impl}

\subsection{Architecture details}

\paragrapht{Static encoder and renderer.}
The static prior used is a pretrained feed-forward novel view synthesizer whose encoder is a $24$-layer transformer alternating frame-wise and global attention, followed by a $12$-layer transformer renderer.
Both use a hidden size of $768$ with $12$ heads on $8\times8$ patches, and remain frozen during our training.

\paragrapht{Dynamic encoder.}
The dynamic encoder adopts the same core architecture as the static encoder, augmented with cross-attention, but uses one-third of its depth, with $8$ layers. It is initialized with weights uniformly sampled from the pretrained static encoder. Each layer additionally cross-attends to the last layer of static features, allowing the residual branch to access camera-pose-aware representations throughout encoding. This provides a spatial anchor for learning motion residuals, complementing the static features rather than relearning scene structure. A DPT decoder~\citep{ranftl2021visiondpt} attached to the dynamic encoder predicts the binary dynamic mask.

\paragrapht{Temporal conditioning.}
We use a single learnable time token $\boldsymbol{\tau} \in \mathbb{R}^{C}$ with $C=1024$, and attach a copy of it to each input view by concatenating it with that view's tokens. 
Each copy becomes view-specific ($\hat{\boldsymbol{\tau}}_k$) through attention with its own view, allowing the same token to handle an arbitrary number and ordering of context frames. The resulting time representation is processed by a time module consisting of two transformer blocks with adaptive layer normalization conditioned on the target-time token, followed by a zero-initialized linear residual gate.

\subsection{Training details}
\label{supp:training}

\paragrapht{Training setup.}
We train with AdamW at a learning rate of $3.5\times10^{-5}$, weight decay $0.05$, and $3$k warmup steps, while keeping the static encoder and renderer frozen for whole training. \ours{} is trained on four NVIDIA RTX 5880 Ada Generation GPUs with a batch size of 8 for total of 32k steps. Each training sample contains 8 conditioning views and 12 target views at a resolution of $336\times512$.

\paragrapht{Training datasets.}
We use Kubric4D~\citep{van2024generativekub4d} and CamxTime~\citep{huang2025spacetimepilotcamxtime}, which provide synchronized multi-view videos captured by static cameras. To disentangle camera and object motion, we train the model to reconstruct target frames at view--time pairs not included in the input set.

\paragrapht{Pose dropout.}
To support pose-free conditioning at inference, we randomly zero out the conditioning camera tokens with probability $0.4$ during training. At inference time, \ours{} supports fully unposed setting by zeroing all conditioning camera tokens.

\paragrapht{Loss.}
We use an $\ell_2$ photometric loss $\mathcal{L}_\mathrm{2}$, LPIPS losses $\mathcal{L}_{\text{perc}}$ and $\mathcal{L}_{\text{perc}}^{\text{dyn}}$, and an $\ell_1$ image-gradient loss $\mathcal{L}_{\text{grad}}^{\text{dyn}}$. The dynamic-region losses are computed using the dynamic mask $M_k$ to select moving pixels, on $128\times128$ crops centered on the moving content. For the image loss term, the dynamic and static regions are averaged over their respective areas and weighted equally. Dynamic mask loss is only comptued for context frames. We set $\lambda_\text{perc}=0.5$, $\lambda_\text{perc}^{\text{dyn}}=2.0$, $\lambda_\text{grad}^{\text{dyn}}=1.0$, and $\lambda_\text{mask}=1.0$.

\paragrapht{Pseudo dynamic mask generation.}
We generate pseudo dynamic masks by thresholding the magnitude of optical flow estimated by an off-the-shelf flow estimator~\citep{wang2026waft}. Since the flow is computed between frames captured by the same static camera, camera-induced motion and parallax are absent, making the remaining flow primarily reflect object motion. This leverages the static-camera multi-view configuration of our training data to identify dynamic regions with binary pseudo labels. The mask generation can also be easily replaced with off-the-shelf segmentation models, such as SAM3~\citep{carion2025sam3segmentconcepts}, without changing the training framework.

\clearpage

\section{Evaluation protocol}
\label{supp:evalproto}

\subsection{Evaluation datasets}
\label{supp:datasets}

Our benchmark comprises five multi-view datasets for NVS and CTVS: NVIDIA, N3DV, Kubric4D, CamxTime, and Charge, and three datasets for NTS: ADT, DyCheck, and TUM. Kubric4D and CamxTime use held-out scenes, while the remaining datasets are unseen during training and evaluate cross-dataset generalization.

For static camera multi-view datasets, we simulate moving camera monocular video by selecting N diagonal frames, \ie camera~$i$ observed at time~$i$ as input.

\paragrapht{Kubric4D.}
We use 100 test scenes from Kubric4D, which contains 16 static cameras, with the first four providing top-down views and the remaining 12 cameras arranged around the scene on a horizontal ring. Each camera captures 60 frames. We exclude the first four cameras and sample cameras consecutively from the horizontal ring. We use a frame interval of 8.

\paragrapht{CamxTime.}
CamxTime contains four separate camera trajectories, each comprising 120 densely spaced cameras with 120 frames per camera. We hold out the last 10\% of scenes for testing and use \texttt{Trajectory\_01} for evaluation, with camera and frame intervals both set to 8.

\paragrapht{Charge.}
we use the \texttt{Sparse\_Test} cameras. Charge stores cameras in a non-spatial order, so sampling by file index leaves large gaps between consecutive conditioning views. We therefore reorder the cameras of each scene along a continuous open path through the rig and uniformly sample along this path to approximate a smooth monocular camera trajectory. we use the following camera order:
\begin{center}
\texttt{C05\ \ C01\ \ C04\ \ C06\ \ C02\ \ C08\ \ C00\ \ C07\ \ C03\ \ C09},
\end{center}
which follows the shortest open path through the rig where \texttt{C\textit{i}} denotes the $i$-th test camera. CTVS uses the first $8$ cameras of this order, and NVS holds out \texttt{C02}. We use a frame interval of 8.

\subsection{Evaluation protocols}
\label{supp:protocols}

\paragrapht{NVS (Novel View Synthesis).}
We take $9$ frames and hold out the $5$th as a target. The remaining $8$ are used as the input, and the held-out frame's viewpoint is the target at all $8$ conditioning times. For N3DV, \texttt{cam0} is used for a target camera.

\paragrapht{CTVS (Cross-time View Synthesis).}
Given $N$ diagonal frames as input, the targets are the $N^2\!-\!N$ off-diagonal cells of the camera--time grid, evaluating synthesis at unseen viewpoint--timestamp combinations. NVIDIA uses all available cameras ($N{=}12$), and N3DV uses all cameras except \texttt{cam0} ($N{=}17$--$20$ depending on the scene). Kubric4D, CamxTime, and Charge each use $N{=}8$.

\paragrapht{NTS (Novel Time Synthesis).}
Following C4G~\citep{c4g}, we use the following scenes from each dataset:
\begin{itemize}
\small
  \item \textbf{DyCheck}
    \begin{itemize}
      \item \texttt{apple}, \texttt{block}, \texttt{paper-windmill}, \texttt{spin}, \texttt{teddy}
    \end{itemize}
  \item \textbf{ADT}
    \begin{itemize}
      \item \texttt{Apartment\_release\_multiuser\_cook\_seq141\_M1292}
      \item \texttt{Apartment\_release\_multiskeleton\_party\_seq114\_M1292}
      \item \texttt{Apartment\_release\_meal\_skeleton\_seq135\_M1292}
      \item \texttt{Apartment\_release\_work\_skeleton\_seq137\_M1292}
    \end{itemize}
  \item \textbf{TUM}
    \begin{itemize}
      \item \texttt{fr2\_desk\_with\_person}
      \item \texttt{fr3\_sitting\_\{halfsphere, rpy, static, xyz\}}
      \item \texttt{fr3\_walking\_\{halfsphere, rpy, static, xyz\}}
    \end{itemize}
\end{itemize}

All three datasets use a single moving camera for both context and target views.
We follow the C4G~\citep{c4g} evaluation protocol and sample 16 frames with stride~6 and evaluate on the 15 interleaved frames.

\subsection{Model specific configurations}
\label{supp:models}

Every model runs at the native input resolution. For the pose-free baselines, NeoVerse~\citep{yang2026neoverse} and C4G~\citep{c4g}, we apply 2{,}000 steps of test-time pose alignment~\citep{ye2025noposplat}. In contrast, \ours{} does not perform pose alignment, despite being evaluated in the pose-free setting. For real-world datasets, \ours{} uses camera poses predicted by VGGT~\citep{wang2025vggt}, using the estimated intrinsics for NVIDIA and the estimated intrinsics and extrinsics for N3DV.

\section{Latent analysis}
\label{supp:feat_analysis}

\input{figures/supp_staticfeat}
\Cref{fig:supp_staticfeat} shows a PCA visualization of the static latent $\mathbf{z}_\text{st}$ under camera motion. The static latent primarily captures camera pose, with features varying smoothly as the camera moves along a circular trajectory. Similar features correspond to similar viewing directions.

\input{figures/supp_nn_residualfeat}
Figure~\ref{fig:supp_nn_residualfeat} visualizes the time-conditioned residual features along with their corresponding GT dynamic masks. For each view $i$ and target time $j$, the residual feature is spatially anchored to the diagonal frame at view $i$ and time $i$. Dynamic objects become more pronounced in the residual as their position at time $j$ increasingly differs from that at the anchor time $i$, while showing weaker responses near the diagonal where the temporal displacement is small. This trend indicates that the residual latent encodes timestamp-dependent differences, with larger feature variations corresponding to larger object motion.

\clearpage

\section{Additional results}
\label{supp:results}

\subsection{Quantitative comparison on dynamic regions}
\label{supp:dynregion}
\input{tables/dynregion}
On the two datasets with ground-truth motion masks, Kubric4D and Charge~\citep{nazarczuk2026charge}, we additionally report metrics computed only within dynamic regions. Since moving content typically occupies a small fraction of the frame, dynamic-region metrics provide a more focused evaluation of reconstruction quality on moving subjects. Table~\ref{tab:dynregion} reports the dynamic-region metrics for NVS and CTVS, where our model consistently outperforms the baselines.

\subsection{Additional analysis on quantitative results}
In the main table, NeoVerse performs notably poorly on Kubric4D. This is because NeoVerse tends to provide weaker information across the temporal axis and produces many holes, leaving regions unseen at a given timestamp largely unmodeled. This issue is particularly severe on Kubric4D, where the large camera gaps span nearly the entire viewing range, making regions observed at different timestamps, such as the background, especially difficult to reconstruct.

\subsection{Additional qualitative results}
\label{supp:qual}
\input{figures/supp_NVS_qual.tex}

Figure~\ref{fig:supp_nvsqual} shows additional qualitative comparisons for 4D NVS. \Cref{fig:supp_baseline_qual1,fig:supp_baseline_qual2,fig:supp_baseline_qual3,fig:supp_baseline_qual4,fig:supp_baseline_qual5,fig:supp_baseline_qual6} provide comparisons across all evaluated methods and scenes.
\Cref{fig:supp_baseline_ctvs_qual1,fig:supp_baseline_ctvs_qual2,fig:supp_baseline_ctvs_qual3} further compare cross-time view synthesis results using view-time grids. Finally, \cref{fig:supp_nvsqual1,fig:supp_nvsqual2,fig:supp_nvsqual3} show additional view-time grid results of \ours{} alongside the corresponding ground-truth views. Across these results, \ours{} exhibits stronger temporal consistency and more effective disentanglement of viewpoint and time, producing stable renderings with fewer temporal inconsistencies and visual artifacts.

\input{tables/dynmask}

\subsection{Dynamic mask}
In Table~\ref{tab:dynmask}, we report the dynamic mask estimation accuracy with IoU and Boundary F-measure as JM and FM, respectively, and their corresponding recall metrics as JR and FR, following the evaluation protocol of~\cite{hu2026vggt4d}. The reported metrics for the other methods are adopted from~\cite{hu2026vggt4d}. Figure~\ref{fig:supp_dynmask} shows that our model predicts dynamic masks with cleaner boundaries and greater robustness, demonstrating its ability to generalize dynamic object localization to unseen real-world datasets.

\input{figures/supp_dynmask}

\subsection{Context frame number ablation}
\label{supp:nframes}
\input{figures/nframe}
Our model is trained with only $8$ context frames, yet Figure~\ref{fig:nframe} shows that the rendering quality remains on a flat plateau as the number of context frames increases to $32+$, more than four times the training configuration. This demonstrates that the dynamic encoder generalizes to substantially longer temporal contexts than trained. The result highlights that our model can effectively exploit temporal context far beyond the number of frames seen during training.

We use CamxTime test scenes, which provide enough number of context frames while maintaining a sufficiently dense set of conditioning candidates. Context frames are sampled along a moving camera trajectory from the camera-time grid within a fixed temporal window. Thus, all settings cover the same camera and temporal extent, while increasing the number of context frames increases the sampling density. We evaluate $12$ target frames across four novel camera viewpoints, ${21, 45, 69, 93}$, and three timestamps, ${8, 56, 104}$.

\subsection{Architectural ablation}
\label{supp:arch_ablation}

\input{tables/supp_ablation}
\Cref{tab:supp_ablation} extends \cref{tab:ablation} to more variants and temporal strides $\{1,2,4,8\}$.
We first compare against the static NVS backbone and its fine-tuned variant. While both provide a strong static-scene prior, fine-tuning alone does not effectively capture scene dynamics.
We further ablate architectural choices. Unfreezing the renderer degrades performance, suggesting that preserving the pretrained latent-to-image mapping is beneficial. The single-branch encoder, removal of the dynamic mask loss, and sinusoidal time embeddings all underperform our full model. Removing static feature anchoring also reduces performance overall, highlighting the benefit of conditioning dynamic features on the static representation.
Finally, we vary the dynamic encoder depth and initialization. Our 8-layer encoder achieves the best overall performance among the tested configurations. Random initialization also yields competitive results, although initializing from the static encoder weights consistently improves performance across temporal strides.

\clearpage

\input{tables/probe_matching}

\input{tables/probe_depth}

\section{Probings}
\label{supp:probing}

\subsection{Full probing results}
Tables~\ref{tab:probe_matching_full} and~\ref{tab:probe_depth_full} provide the full probing results, while the main paper reports a subset of matching thresholds and depth across temporal strides.

\subsection{Point tracking}
\input{tables/probe_tracking}
\input{figures/supp_tracking_qual}
Correspondence between frames does not necessarily capture how scene points evolve over time. We therefore evaluate point tracking by attaching a CoTracker3~\citep{karaev2025cotracker3} head to each representation. All variants are trained on Kubric~\citep{greff2022kubric} for the same number of epochs and evaluated in DAVIS~\citep{pont20172017davis} following the protocol of~\citep{son2025probingditracker}. \Cref{fig:supp_trackingqual} provides qualitative comparisons, showing that dynamic features yield more accurate point tracks over time. \Cref{tab:probe_tracking_full} shows that dynamic feature substantially improves tracking of moving points over the static feature, with the performance gap increasing as the temporal stride grows. Consistent with the depth probing results, this widening gap indicates that the residual captures temporal information that is particularly useful under larger object displacement.

\subsection{Video Depth estimation}
\input{figures/supp_probe_depth_qual}

\Cref{fig:supp_depthqual} further compares depth maps predicted from static and dynamic features. Dynamic features yield more accurate depth predictions, with sharper boundaries, more coherent depth within objects, and better background depth estimation, whereas static features produce noisier depth maps.

\section{Test-time pose alignment for latent models}
\label{supp:tto}

Evaluating models that do not receive conditioning poses can introduce scale misalignment with the ground truth, resulting in poor rendering metrics even when the predicted scene is geometrically accurate. Since the input images do not determine an absolute scale, a pose-free model may reconstruct the scene at an arbitrary scale. Pose-free methods therefore perform test-time pose alignment before evaluation and report rendering metrics using the aligned poses~\citep{ye2025noposplat, c4g}. This alignment only brings the prediction into the coordinate frame and scale of the ground truth without modifying the underlying scene content.

The pose-free Gaussian baselines in our quantitative comparisons are evaluated with 2{,}000 step pose-alignment steps, whereas \ours{} is reported without pose alignment in the main tables. However, \ours{} is also pose-free and therefore suffers from the same scale ambiguity that motivates this evaluation procedure. We thus additionally evaluate \ours{} with the same 2{,}000  step pose alignment to provide a consistent comparison.

For Gaussian models, Gaussians are predicted once from the conditioning images and kept fixed throughout the alignment process. Differentiable rasterization allows the rendering loss to be backpropagated to the camera pose, so each alignment step re-rasterizes the same frozen Gaussians under an updated pose, without modifying the network or scene representation.

The analogous procedure can be done to our latent renderer. The scene latent is predicted solely from the conditioning images and is independent of the rendering process. We therefore compute the latent once and keep it fixed throughout the alignment process, with the renderer serving the role of rasterization by re-rendering the cached latent under an updated target pose at each step. For both Gaussian and latent models, we optimize a rotation and translation offset for each target pose~\citep{ye2025noposplat}, initialized at zero such that the initial rendering exactly matches the feed-forward prediction. We use the same objective as the Gaussian baselines, with 2{,}000 Adam steps using pixel and perceptual losses. Table~\ref{tab:tto} reports the effect of this alignment under the NTS protocol on ADT and TUM. Alignment is worth $5.94$ dB to C4G on ADT and $5.13$ dB on TUM, against $3.59$ dB and $2.55$ dB for our model, so a large part of the published margin of C4G on these cells comes from fitting the pose to the image it is then scored on. How much alignment recovers therefore measures how much error the supplied cameras carry rather than the quality of the reconstruction, and our model without any alignment remains ahead of both aligned baselines on every metric. \input{tables/tto}

\section{Limitations}
\label{supp:limitations}
Our approach inherits the limitations of the pretrained static latent NVS backbone, particularly when rendering from extrapolated viewpoints that are unseen in the input views. Latent NVS models cannot synthesize unseen content~\citep{lvsm}, resulting in blurry or collapsed renderings under extrapolation, as also observed in \citep{szymanowicz2026lagernvs}. In contrast, explicit Gaussian representations construct scene content in 3D space, allowing the observed portions of the scene to remain usable even when extrapolation leaves holes in unseen regions.

The residual formulation can compensate for certain errors in the static prediction, but its capacity is naturally limited when the underlying scene representation or camera pose is substantially incorrect. These limitations can be alleviated by using stronger static NVS models with broader scene coverage and more robust viewpoint understanding.

\clearpage

\input{figures/supp_baseline_qual2} 
\input{figures/supp_baseline_qual3} 
\input{figures/supp_baseline_qual6} 
\input{figures/supp_baseline_qual1} 
\input{figures/supp_baseline_qual4} 
\input{figures/supp_baseline_qual5} 

\input{figures/supp_baseline_ctvs_qual_1} 
\input{figures/supp_baseline_ctvs_qual_3} 
\input{figures/supp_baseline_ctvs_qual_2} 

\input{figures/supp_nvs_qual1.tex}
\input{figures/supp_nvs_qual2.tex}
\input{figures/supp_nvs_qual3.tex}

%% file: figures/supp_staticfeat.tex
\begin{figure}[t]
\centering

\includegraphics[width=1.00\textwidth]{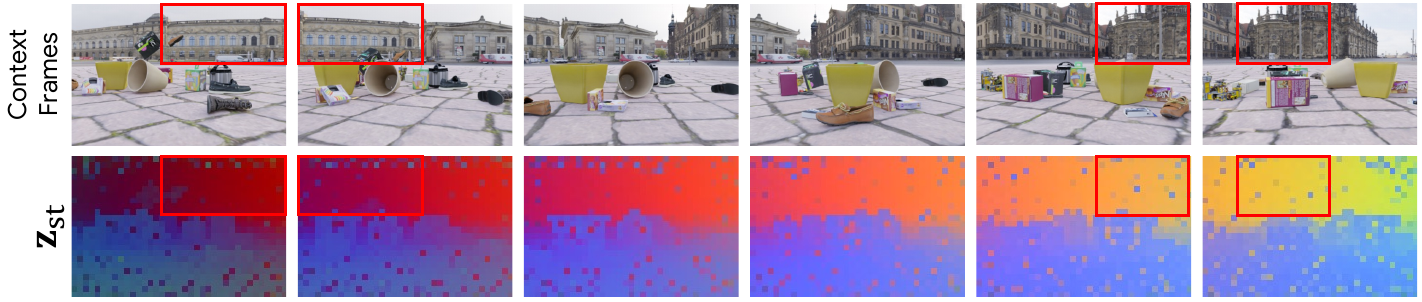}

\caption{\textbf{Visualization of static latent.} The static latent encodes camera pose, with similar colors for cameras pointing in the same direction. Red boxes highlight overlapping regions across context frames, which exhibit similar color gradients.}
\label{fig:supp_staticfeat}
\end{figure}

%% file: figures/supp_nn_residualfeat.tex
\begin{figure}[t]
\centering

\includegraphics[width=1.00\textwidth]{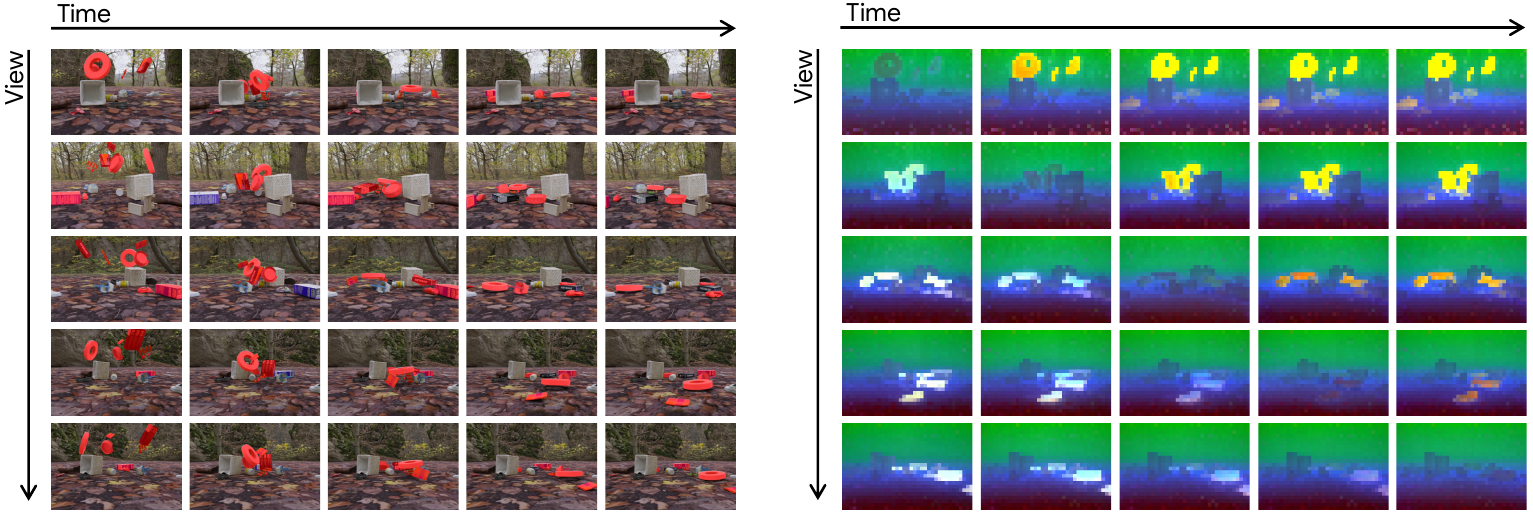}

\caption{\textbf{Visualization of residual latent.} \figleft{} Ground-truth RGB frame overlaid with the dynamic mask. \figright{} Corresponding residual latent grid. Dynamic objects exhibit more pronounced residual latents as the target pose and time diverge from the input, indicating a larger discrepancy from the context frames.}
\label{fig:supp_nn_residualfeat}
\end{figure}

%% file: tables/dynregion.tex
\begin{table}[!t]
\centering
\caption{\textbf{4D NVS result for Dynamic region.} $^\dagger$ denotes 2{,}000 step test-time pose alignment, the \firstbox{best}, \secondbox{second best}, and \thirdbox{third best} results in each column are highlighted. Metrics are restricted to the moving pixels and are computed over the same target subset, and their full-frame counterparts for the same cells are reported in \cref{tab:nvs,tab:ctvs}.}
\label{tab:dynregion}
\resizebox{\textwidth}{!}{%
\begin{tabular}{@{}l c ccc ccc ccc ccc@{}}
\toprule
\multirow{3}{*}{Method} & \multirow{3}{*}{Pose}
& \multicolumn{6}{c}{NVS} & \multicolumn{6}{c}{CTVS} \\
\cmidrule(lr){3-8}\cmidrule(lr){9-14}
& & \multicolumn{3}{c}{Kubric4D} & \multicolumn{3}{c}{Charge}
  & \multicolumn{3}{c}{Kubric4D} & \multicolumn{3}{c}{Charge} \\
\cmidrule(lr){3-5}\cmidrule(lr){6-8}\cmidrule(lr){9-11}\cmidrule(lr){12-14}
& & PSNR$\uparrow$ & SSIM$\uparrow$ & LPIPS$\downarrow$ & PSNR$\uparrow$ & SSIM$\uparrow$ & LPIPS$\downarrow$
  & PSNR$\uparrow$ & SSIM$\uparrow$ & LPIPS$\downarrow$ & PSNR$\uparrow$ & SSIM$\uparrow$ & LPIPS$\downarrow$ \\
\midrule
\multicolumn{8}{@{}l}{\emph{Feed-forward 4D Gaussian Representation}}\\
4DGT~\citep{4dgt} & \cmark & 12.089 & 0.2406 & \cellcolor{tabthird}0.0305 & 17.703 & 0.6516 & \cellcolor{tabthird}0.1127 & 12.405 & 0.2557 & \cellcolor{tabthird}0.0303 & 17.708 & \cellcolor{tabthird}0.6588 & 0.1157 \\
MoVieS~\citep{lin2026movies} & \cmark & 11.401 & 0.1741 & 0.0318 & 17.969 & 0.6079 & 0.1130 & 11.797 & 0.2034 & 0.0307 & 18.068 & 0.6119 & \cellcolor{tabthird}0.1147 \\
NeoVerse$^\dagger$~\citep{yang2026neoverse} & \xmark & 10.847 & 0.1527 & 0.0381 & 16.292 & 0.5161 & 0.1242 & 10.254 & 0.1372 & 0.0403 & 16.152 & 0.4966 & 0.1307 \\
C4G$^\dagger$~\citep{c4g} & \xmark & \cellcolor{tabthird}13.031 & \cellcolor{tabthird}0.3082 & 0.0322 & \cellcolor{tabthird}18.552 & \cellcolor{tabthird}0.6617 & 0.1143 & \cellcolor{tabthird}12.901 & \cellcolor{tabthird}0.3052 & 0.0333 & \cellcolor{tabthird}18.385 & 0.6566 & 0.1166 \\
\midrule
\multicolumn{8}{@{}l}{\emph{Feed-forward Latent-based 4D Representation}}\\
\textbf{\ours{} (Ours)} & \xmark & \cellcolor{tabsecond}14.774 & \cellcolor{tabsecond}0.3479 & \cellcolor{tabsecond}0.0240 & \cellcolor{tabsecond}19.906 & \cellcolor{tabsecond}0.6708 & \cellcolor{tabsecond}0.1063 & \cellcolor{tabfirst}14.482 & \cellcolor{tabsecond}0.3351 & \cellcolor{tabsecond}0.0246 & \cellcolor{tabsecond}19.748 & \cellcolor{tabfirst}0.6774 & \cellcolor{tabsecond}0.1019 \\
\textbf{\ours{} (Ours)} & \cmark & \cellcolor{tabfirst}14.800 & \cellcolor{tabfirst}0.3485 & \cellcolor{tabfirst}0.0236 & \cellcolor{tabfirst}19.993 & \cellcolor{tabfirst}0.6711 & \cellcolor{tabfirst}0.1029 & \cellcolor{tabsecond}14.451 & \cellcolor{tabfirst}0.3358 & \cellcolor{tabfirst}0.0244 & \cellcolor{tabfirst}19.935 & \cellcolor{tabsecond}0.6772 & \cellcolor{tabfirst}0.1009 \\
\bottomrule
\end{tabular}%
}
\end{table}

%% file: figures/supp_NVS_qual.tex
\begin{figure}[t]
\centering

\includegraphics[width=1\textwidth]{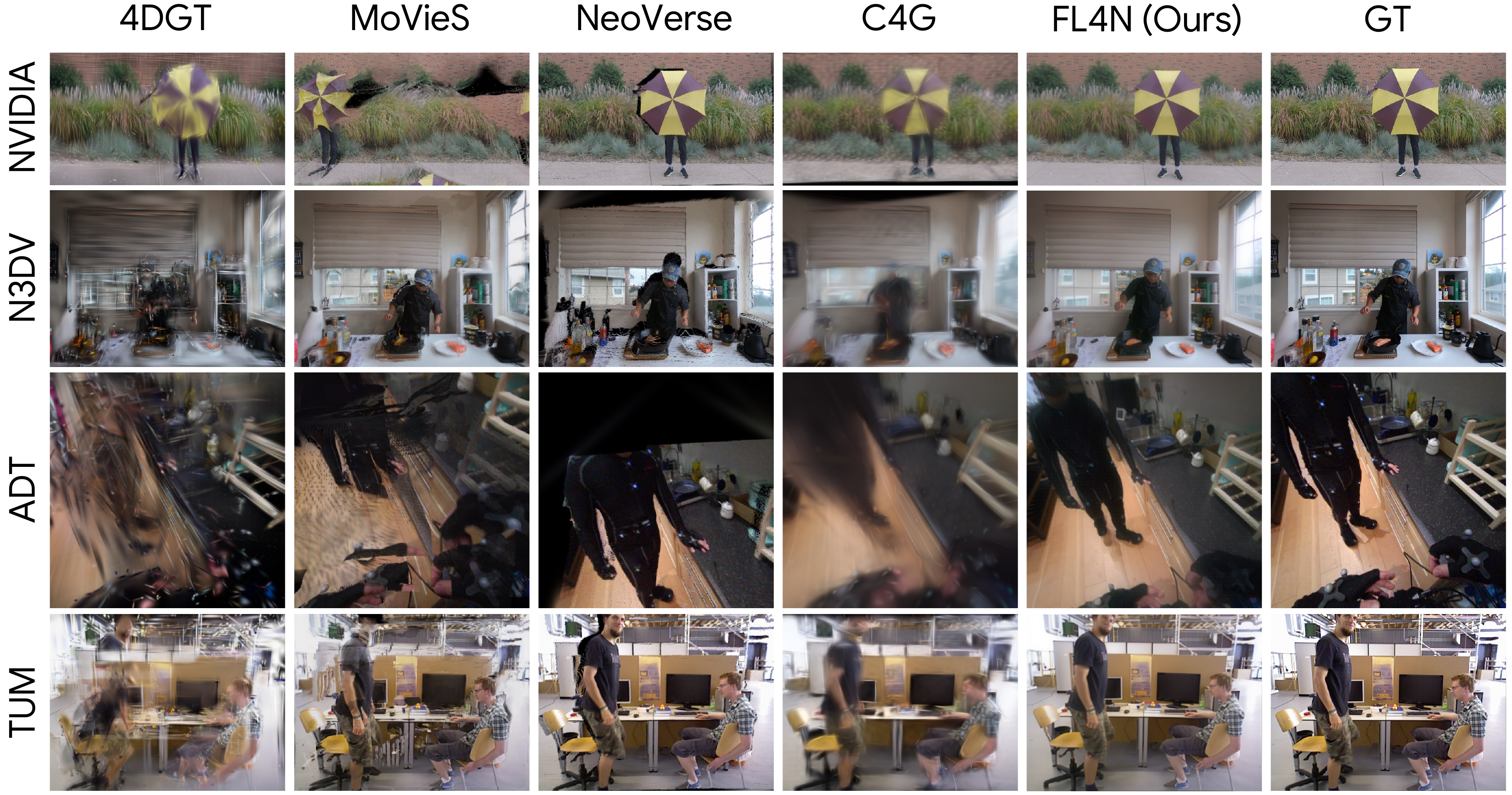}

\caption{\textbf{Further qualitative comparison.}
Gaussian-based methods exhibit holes, object inconsistencies, and blurred details. Our method produces more coherent and visually consistent results, demonstrating the superiority of our latent-based representation over explicit geometry primitives.
}
\label{fig:supp_nvsqual}
\end{figure}

%% file: tables/dynmask.tex
\begin{table}[t]
\centering
\caption{\textbf{Quantitative results on dynamic regions.}
On the dynamic object segmentation task, Our method substantially outperforms other methods for 3D or 4D geometry estimation.
\firstbox{Best} and \secondbox{second best} results are highlighted.
}
\label{tab:dynmask}
\scriptsize
\setlength\tabcolsep{4.5pt}

\begin{tabularx}{\columnwidth}{@{}X cccc cccc cccc@{}}
\toprule

\multirow{2}{*}{Method}
& \multicolumn{4}{c}{DAVIS-2016} & \multicolumn{4}{c}{DAVIS-2017} & \multicolumn{4}{c}{DAVIS-all} \\
\cmidrule(lr){2-5} \cmidrule(lr){6-9} \cmidrule(lr){10-13}
 & JM$\uparrow$ & JR$\uparrow$ & FM$\uparrow$ & FR$\uparrow$ & JM$\uparrow$ & JR$\uparrow$ & FM$\uparrow$ & FR$\uparrow$ &JM$\uparrow$ & JR$\uparrow$ & FM$\uparrow$ & FR$\uparrow$ \\
\midrule
Easi3R$_\text{dust3r}$ \citep{chen2025easi3r} & 50.10 & 55.77 & 43.40 & 37.25 & 46.86 & 50.54 & 39.06 & 30.05 & 44.10 & 50.85 & 35.16 & 27.24 \\
Easi3R$_\text{monst3r}$ \citep{chen2025easi3r} & 54.93 & 68.00 & 45.29 & 47.30 & 54.75 & \cellcolor{tabsecond}66.16 & 44.09 & 42.36 & \cellcolor{tabsecond}51.64 & \cellcolor{tabsecond}63.06 & 40.98 & 38.49 \\
MonST3R \citep{ICLR2025_ce1710ff} & 40.42 & 40.39 & 49.54 & 52.12 & 38.07 & 36.05 & 48.24 & 49.01 & 36.98 & 34.52 & 47.03 & \cellcolor{tabsecond}46.72 \\
DAS3R \citep{xu2024das3r} & 41.13 & 38.67 & 44.50 & 36.94 & 44.51 & 43.95 & 46.71 & 44.96 & 43.33 & 38.93 & 45.24 & 38.78 \\
VGGT4D \citep{hu2026vggt4d} & \cellcolor{tabsecond}62.12 & \cellcolor{tabsecond}76.80 & \cellcolor{tabsecond}56.04 & \cellcolor{tabsecond}67.49 & \cellcolor{tabsecond}56.45 & 65.62 & \cellcolor{tabsecond}51.09 & \cellcolor{tabsecond}56.85 & 50.75 & 55.59 & \cellcolor{tabsecond}47.04 & 46.43 \\
\midrule
\textbf{\ours{} (Ours)} & \cellcolor{tabfirst}74.93 & \cellcolor{tabfirst}93.78 & \cellcolor{tabfirst}74.16 & \cellcolor{tabfirst}90.64 & \cellcolor{tabfirst}75.44 & \cellcolor{tabfirst}92.32 & \cellcolor{tabfirst}74.37 & \cellcolor{tabfirst}90.38 & \cellcolor{tabfirst}68.76 & \cellcolor{tabfirst}80.11 & \cellcolor{tabfirst}69.87 & \cellcolor{tabfirst}83.99 \\

\bottomrule
\end{tabularx}%

\end{table}

%% file: figures/supp_dynmask.tex
\begin{figure}[t]
\centering

\includegraphics[width=\textwidth]{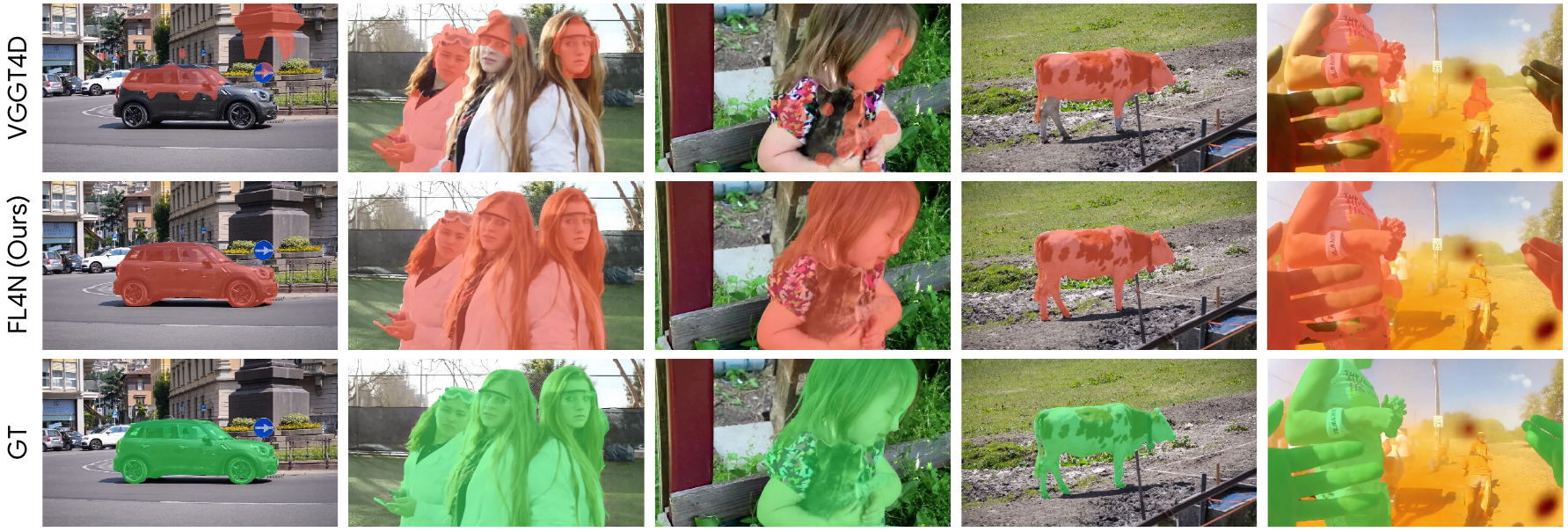}

\caption{\textbf{Qualitative comparison of dynamic masks.} Our method show more accurate dynamic object masks with clearer object boundary than VGGT4D \citep{hu2026vggt4d}.}
\label{fig:supp_dynmask}
\end{figure}

%% file: figures/nframe.tex
\begin{figure}[t]
\centering
\includegraphics[width=0.62\textwidth]{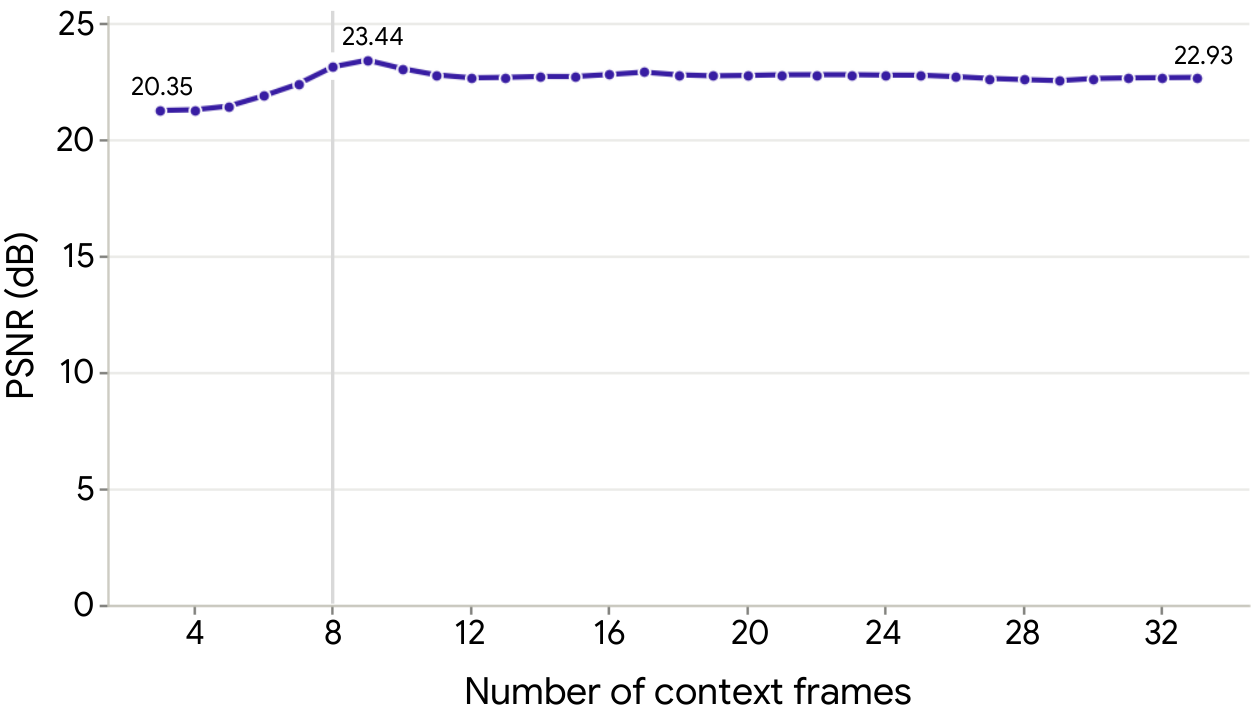}

\caption{\textbf{Generalization to the number of context frames}, reporting PSNR on CamxTime as the number of conditioning frames varies at inference. The vertical line marks the $8$ frames the model is trained with.
The rendering quality remains on a flat plateau as the number of context frames increases to $32+$.
The dynamic encoder generalizes to substantially longer temporal contexts than trained and effectively exploits temporal context far beyond the number of frames seen during training.
}
\label{fig:nframe}
\end{figure}

%% file: tables/supp_ablation.tex
\begin{table}[!t]
\centering
\caption{\textbf{Extensive ablation on Kubric4D} across different temporal strides. \firstbox{Best} and \secondbox{second best} results are highlighted.}
\label{tab:supp_ablation}
\scriptsize
\setlength\tabcolsep{3pt}
\resizebox{\textwidth}{!}{%
\begin{tabular}{@{}l ccc ccc ccc ccc@{}}
\toprule
\multirow{2}{*}{Variant}
& \multicolumn{3}{c}{$\Delta$1}
& \multicolumn{3}{c}{$\Delta$2}
& \multicolumn{3}{c}{$\Delta$4}
& \multicolumn{3}{c}{$\Delta$8} \\
\cmidrule(lr){2-4}
\cmidrule(lr){5-7}
\cmidrule(lr){8-10}
\cmidrule(lr){11-13}
& PSNR$\uparrow$ & SSIM$\uparrow$ & LPIPS$\downarrow$
 & PSNR$\uparrow$ & SSIM$\uparrow$ & LPIPS$\downarrow$
 & PSNR$\uparrow$ & SSIM$\uparrow$ & LPIPS$\downarrow$
 & PSNR$\uparrow$ & SSIM$\uparrow$ & LPIPS$\downarrow$ \\
\midrule
\multicolumn{13}{l}{\emph{Baselines}} \\
Static NVS model
& 21.90 & 0.727 & 0.287 & 20.87 & 0.716 & 0.308 & 20.49 & 0.705 & 0.325 & 20.35 & 0.700 & 0.332 \\
Static NVS model, fine-tuned
& 21.83 & 0.668 & 0.320 & 21.30 & 0.664 & 0.334 & 21.32 & 0.666 & 0.338 & 21.33 & 0.665 & 0.339 \\
\midrule
\multicolumn{13}{l}{\emph{Architecture}} \\
Unfrozen renderer
& 22.15 & 0.681 & 0.309 & 21.57 & 0.676 & 0.325 & 21.55 & 0.676 & 0.332 & 21.55 & 0.676 & 0.334 \\
Single branch encoder
& 23.18 & 0.727 & 0.290 & 22.74 & 0.722 & 0.300 & 22.58 & 0.718 & 0.307 & 22.47 & 0.715 & 0.312 \\
w/o dynamic mask loss
& 22.93 & 0.741 & 0.262 & 22.24 & 0.728 & 0.285 & 22.05 & 0.720 & 0.298 & 21.99 & 0.719 & 0.303 \\
w/o static feature anchoring
& 23.61 & 0.769 & \cellcolor{tabsecond}0.246 & \cellcolor{tabfirst}22.91 & 0.755 & \cellcolor{tabsecond}0.267 & \cellcolor{tabsecond}22.61 & 0.744 & \cellcolor{tabsecond}0.282 & \cellcolor{tabsecond}22.47 & 0.741 & 0.289 \\
Sinusoidal time embedding
& 23.02 & 0.747 & 0.259 & 22.29 & 0.730 & 0.279 & 22.19 & 0.715 & 0.295 & 22.12 & 0.715 & 0.301 \\
\midrule
\multicolumn{13}{l}{\emph{Dynamic encoder}} \\
4 layers
& 23.58 & 0.771 & 0.247 & 22.76 & \cellcolor{tabsecond}0.756 & 0.269 & 22.52 & \cellcolor{tabsecond}0.745 & 0.283 & 22.44 & \cellcolor{tabsecond}0.743 & \cellcolor{tabsecond}0.289 \\
16 layers
& \cellcolor{tabsecond}23.63 & 0.769 & 0.249 & 22.87 & 0.754 & 0.271 & 22.57 & 0.744 & 0.286 & 22.45 & 0.740 & 0.292 \\
24 layers
& 23.44 & 0.755 & 0.256 & 22.67 & 0.740 & 0.278 & 22.37 & 0.730 & 0.292 & 22.24 & 0.726 & 0.299 \\
8 layers, random init.
& 23.42 & \cellcolor{tabsecond}0.772 & 0.250 & 22.56 & 0.755 & 0.275 & 22.32 & 0.743 & 0.290 & 22.15 & 0.739 & 0.296 \\
\midrule
\textbf{\ours{} (Ours)}
& \cellcolor{tabfirst}23.67 & \cellcolor{tabfirst}0.773 & \cellcolor{tabfirst}0.237 & \cellcolor{tabsecond}22.88 & \cellcolor{tabfirst}0.760 & \cellcolor{tabfirst}0.257 & \cellcolor{tabfirst}22.65 & \cellcolor{tabfirst}0.751 & \cellcolor{tabfirst}0.271 & \cellcolor{tabfirst}22.52 & \cellcolor{tabfirst}0.747 & \cellcolor{tabfirst}0.278 \\
\bottomrule
\end{tabular}%
}
\end{table}

%% file: tables/probe_matching.tex
\begin{table}[t]
\centering
\caption{\textbf{Zero-shot matching on DAVIS.} Performance on all threshold are reported.
Dynamic latent demonstrates consistently better zero-shot matching accuracy than static latent on all thresholds.}
\label{tab:probe_matching_full}

\scriptsize
\setlength\tabcolsep{4.2pt}

\begin{tabularx}{0.66\columnwidth}{@{}XX c c c c c c@{}}
\toprule
Region
& Feat.
& $\delta_\text{avg}$ & $\delta$1 & $\delta$2 & $\delta$4 & $\delta$8 & $\delta$16 \\
\midrule
\multirow{2}{*}{All}
& $\mathbf{z}_\text{st}$ & 27.33 & 5.65 & 12.43 & 25.37 & 39.78 & 53.40 \\
& $\mathbf{z}_\text{dyn}$ & \textbf{29.04} & \textbf{5.76} & \textbf{12.50}  & \textbf{26.50} & \textbf{42.81} & \textbf{57.64}  \\
&& {\tiny\textcolor{mygreen}{($+$1.71)}} 
    &  {\tiny\textcolor{mygreen}{($+$0.11)}} 
    &  {\tiny\textcolor{mygreen}{($+$0.07)}} 
    &  {\tiny\textcolor{mygreen}{($+$1.13)}} 
    & {\tiny\textcolor{mygreen}{($+$3.03)}} 
    & {\tiny\textcolor{mygreen}{($+$4.24)}} \\
\midrule
\multirow{2}{*}{Dyn}
& $\mathbf{z}_\text{st}$ & 23.45 & 3.10 & 9.12 & 21.14 & 34.84 & 49.03 \\
& $\mathbf{z}_\text{dyn}$ & \textbf{25.29} & \textbf{3.12} & \textbf{9.20} & \textbf{22.37} & \textbf{38.05} & \textbf{53.68} \\
&& {\tiny\textcolor{mygreen}{($+$1.84)}} 
    &  {\tiny\textcolor{mygreen}{($+$0.02)}} 
    &  {\tiny\textcolor{mygreen}{($+$0.08)}} 
    &  {\tiny\textcolor{mygreen}{($+$1.23)}} 
    & {\tiny\textcolor{mygreen}{($+$3.21)}} 
    & {\tiny\textcolor{mygreen}{($+$4.65)}} \\
\bottomrule
\end{tabularx}%
\end{table}

%% file: tables/probe_depth.tex
\begin{table}[!t]
\centering
\caption{\textbf{Video depth probed on Kubric.} Performance on all strides are reported. The gain is larger at larger temporal gap.}
\label{tab:probe_depth_full}

\scriptsize
\setlength\tabcolsep{4.2pt}

\begin{tabularx}{\columnwidth}{@{}XX cc cc cc cc@{}}
\toprule
\multirow{2}{*}{Region}
& \multirow{2}{*}{Feat.}
& \multicolumn{2}{c}{$\Delta$1}
& \multicolumn{2}{c}{$\Delta$2}
& \multicolumn{2}{c}{$\Delta$4}
& \multicolumn{2}{c}{$\Delta$8} \\
\cmidrule(lr){3-4}
\cmidrule(lr){5-6}
\cmidrule(lr){7-8}
\cmidrule(lr){9-10}
&& AbsRel$\downarrow$ & $\delta<1.25\uparrow$
 & AbsRel$\downarrow$ & $\delta<1.25\uparrow$
 & AbsRel$\downarrow$ & $\delta<1.25\uparrow$
 & AbsRel$\downarrow$ & $\delta<1.25\uparrow$ \\
\midrule

\multirow{3}{*}{All}
& $\mathbf{z}_\text{st}$
& 0.1129 & 0.8550
& 0.1023 & 0.8729
& 0.0944 & 0.8908
& 0.0868 & 0.9081 \\
& $\mathbf{z}_\text{dyn}$
& \textbf{0.1037} & \textbf{0.8722}
& \textbf{0.0932} & \textbf{0.8937}
& \textbf{0.0863} & \textbf{0.9035}
& \textbf{0.0751} & \textbf{0.9326} \\
&& {\scriptsize\textcolor{mygreen}{($-$0.0092)}}
 & {\scriptsize\textcolor{mygreen}{($+$0.0171)}}
 & {\scriptsize\textcolor{mygreen}{($-$0.0092)}}
 & {\scriptsize\textcolor{mygreen}{($+$0.0208)}}
 & {\scriptsize\textcolor{mygreen}{($-$0.0081)}}
 & {\scriptsize\textcolor{mygreen}{($+$0.0127)}}
 & {\scriptsize\textcolor{mygreen}{($-$0.0117)}}
 & {\scriptsize\textcolor{mygreen}{($+$0.0244)}} \\
\midrule

\multirow{3}{*}{Dyn}
& $\mathbf{z}_\text{st}$
& 0.1180 & 0.8597
& 0.1194 & 0.8614
& 0.1178 & 0.8699
& 0.1149 & 0.8790 \\
& $\mathbf{z}_\text{dyn}$
& \textbf{0.1127} & \textbf{0.8747}
& \textbf{0.1078} & \textbf{0.8824}
& \textbf{0.1050} & \textbf{0.8941}
& \textbf{0.0996} & \textbf{0.9099} \\
&& {\scriptsize\textcolor{mygreen}{($-$0.0052)}}
 & {\scriptsize\textcolor{mygreen}{($+$0.0150)}}
 & {\scriptsize\textcolor{mygreen}{($-$0.0116)}}
 & {\scriptsize\textcolor{mygreen}{($+$0.0209)}}
 & {\scriptsize\textcolor{mygreen}{($-$0.0128)}}
 & {\scriptsize\textcolor{mygreen}{($+$0.0242)}}
 & {\scriptsize\textcolor{mygreen}{($-$0.0153)}}
 & {\scriptsize\textcolor{mygreen}{($+$0.0309)}} \\
\bottomrule
\end{tabularx}%
\end{table}

%% file: tables/probe_tracking.tex
\begin{table}[!t]
\centering
\caption{\textbf{Point tracking probed on DAVIS.} Tracking performance across different temporal strides are reported.}
\label{tab:probe_tracking_full}

\scriptsize
\setlength\tabcolsep{1pt}

\begin{tabularx}{\columnwidth}{@{}
X @{\hspace{2pt}}
l @{\hspace{5pt}}
ccc@{}p{5pt}@{}
ccc@{}p{5pt}@{}
ccc@{}p{5pt}@{}
ccc@{}}
\toprule
\multirow{2}{*}{Region}
& \multirow{2}{*}{Feat.}
& \multicolumn{3}{c}{$\Delta$1}
&& \multicolumn{3}{c}{$\Delta$2}
&& \multicolumn{3}{c}{$\Delta$4}
&& \multicolumn{3}{c}{$\Delta$8} \\
\cmidrule(lr){3-5}
\cmidrule(lr){7-9}
\cmidrule(lr){11-13}
\cmidrule(lr){15-17}
&& AJ$\uparrow$ & $<\delta_{\mathrm{avg}}^{x}\uparrow$ & OA$\uparrow$
&& AJ$\uparrow$ & $<\delta_{\mathrm{avg}}^{x}\uparrow$ & OA$\uparrow$
&& AJ$\uparrow$ & $<\delta_{\mathrm{avg}}^{x}\uparrow$ & OA$\uparrow$
&& AJ$\uparrow$ & $<\delta_{\mathrm{avg}}^{x}\uparrow$ & OA$\uparrow$ \\
\midrule
\multirow{3}{*}{All}
& $\mathbf{z}_\text{st}$
& 37.63 & 52.35 & 70.40
&& 36.15 & 51.65 & 69.20
&& 33.10 & 49.00 & 66.42
&& 28.92 & 45.82 & 63.42 \\
& $\mathbf{z}_\text{dyn}$
& \textbf{40.45} & \textbf{54.84} & \textbf{73.25}
&& \textbf{39.80} & \textbf{54.76} & \textbf{72.61}
&& \textbf{36.59} & \textbf{51.73} & \textbf{69.79}
&& \textbf{32.29} & \textbf{47.60} & \textbf{67.06} \\
&& {\scriptsize\textcolor{mygreen}{($+$2.82)}}
& {\scriptsize\textcolor{mygreen}{($+$2.49)}}
& {\scriptsize\textcolor{mygreen}{($+$2.85)}}
&& {\scriptsize\textcolor{mygreen}{($+$3.64)}}
& {\scriptsize\textcolor{mygreen}{($+$3.11)}}
& {\scriptsize\textcolor{mygreen}{($+$3.41)}}
&& {\scriptsize\textcolor{mygreen}{($+$3.49)}}
& {\scriptsize\textcolor{mygreen}{($+$2.73)}}
& {\scriptsize\textcolor{mygreen}{($+$3.37)}}
&& {\scriptsize\textcolor{mygreen}{($+$3.38)}}
& {\scriptsize\textcolor{mygreen}{($+$1.78)}}
& {\scriptsize\textcolor{mygreen}{($+$3.64)}} \\
\midrule
\multirow{3}{*}{Dyn}
& $\mathbf{z}_\text{st}$
& 32.26 & 47.85 & 66.12
&& 30.83 & 47.39 & 64.80
&& 28.57 & 45.69 & 62.65
&& 24.84 & 42.88 & 59.72 \\
& $\mathbf{z}_\text{dyn}$
& \textbf{34.82} & \textbf{49.86} & \textbf{69.23}
&& \textbf{34.09} & \textbf{49.53} & \textbf{68.65}
&& \textbf{32.01} & \textbf{47.70} & \textbf{66.64}
&& \textbf{28.75} & \textbf{44.81} & \textbf{64.38} \\
&& {\scriptsize\textcolor{mygreen}{($+$2.56)}}
& {\scriptsize\textcolor{mygreen}{($+$2.02)}}
& {\scriptsize\textcolor{mygreen}{($+$3.11)}}
&& {\scriptsize\textcolor{mygreen}{($+$3.27)}}
& {\scriptsize\textcolor{mygreen}{($+$2.14)}}
& {\scriptsize\textcolor{mygreen}{($+$3.85)}}
&& {\scriptsize\textcolor{mygreen}{($+$3.43)}}
& {\scriptsize\textcolor{mygreen}{($+$2.00)}}
& {\scriptsize\textcolor{mygreen}{($+$3.99)}}
&& {\scriptsize\textcolor{mygreen}{($+$3.91)}}
& {\scriptsize\textcolor{mygreen}{($+$1.93)}}
& {\scriptsize\textcolor{mygreen}{($+$4.65)}} \\
\bottomrule
\end{tabularx}%
\end{table}

%% file: figures/supp_tracking_qual.tex
\begin{figure}[t]
\centering

\includegraphics[width=1\textwidth]{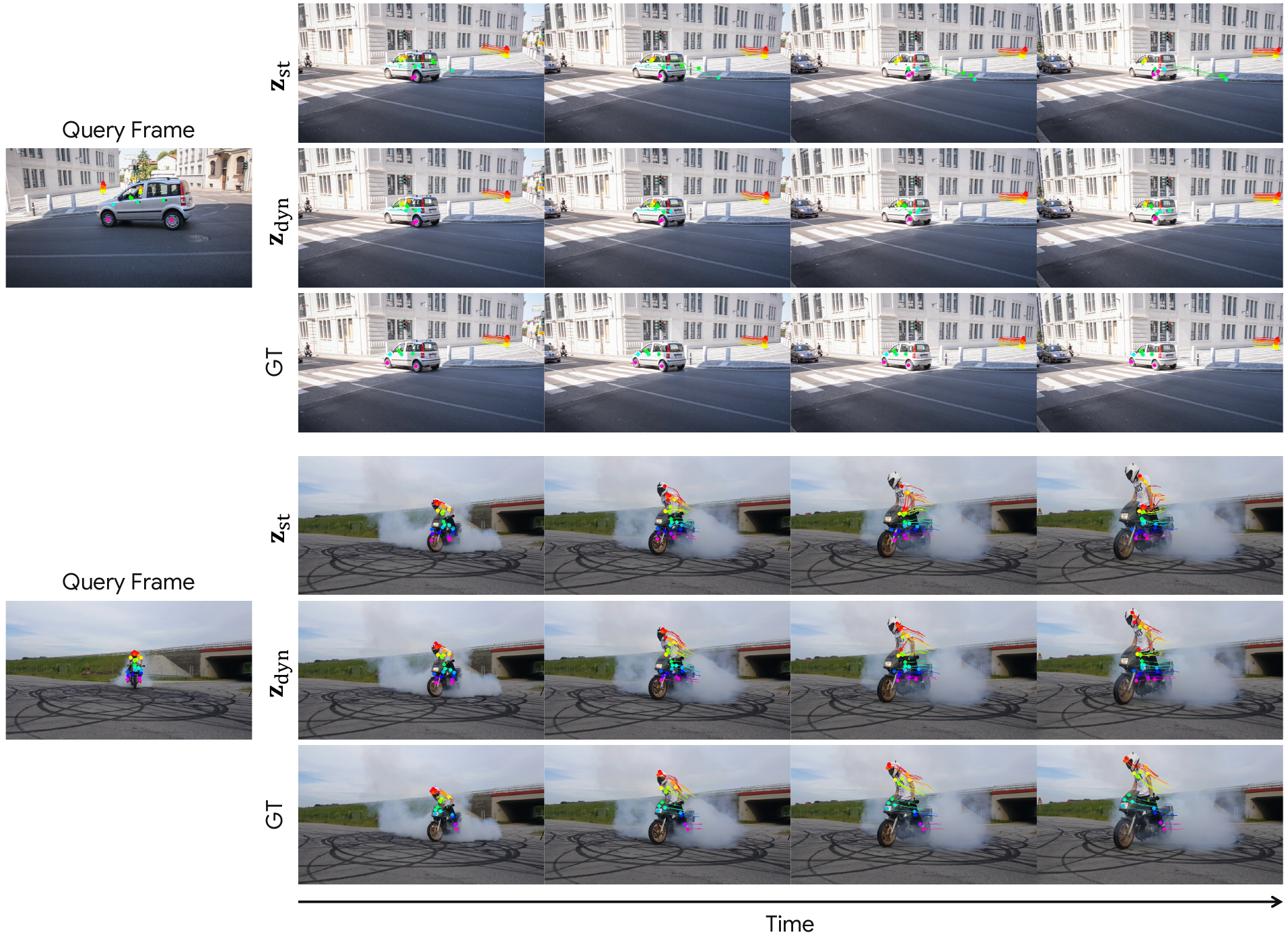}

\caption{\textbf{Qualitative comparison of point tracking from static and dynamic features.}
Compared to static features, dynamic features produce more temporally coherent point tracks by better following objects and distinguishing them from the background.
}
\label{fig:supp_trackingqual}
\end{figure}

%% file: figures/supp_probe_depth_qual.tex
\begin{figure}[t]
\centering

\includegraphics[width=1\textwidth]{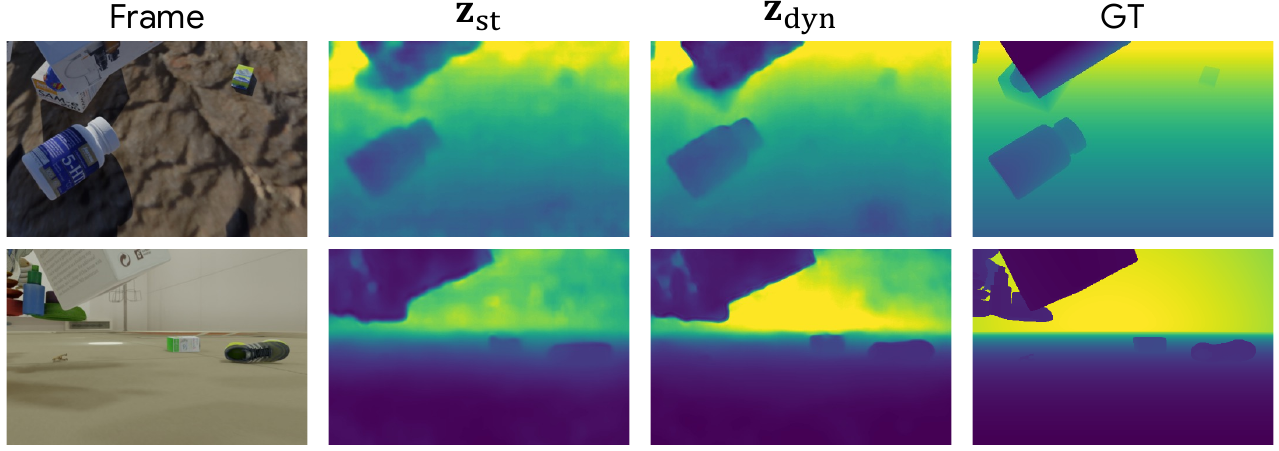}

\caption{\textbf{Qualitative comparison of depth map prediction from static and dynamic features.}
Compared to static features, dynamic features produce clearer depth maps, with more coherent depth predictions across objects and the background.
}
\label{fig:supp_depthqual}
\end{figure}

%% file: tables/tto.tex
\begin{table}[t]
\centering
\caption{\textbf{Test-time pose-alignment results for pose-free methods.} Alignment optimizes each target pose against the ground-truth target image for 2{,}000 steps, and the green row reports its effect. \firstbox{Best}, \secondbox{second best}, and \thirdbox{third best} results are highlighted.}
\label{tab:tto}
\scriptsize
\setlength\tabcolsep{5pt}

\begin{tabularx}{0.9\textwidth}{@{}X c ccc ccc@{}}
\toprule
\multirow{2}{*}{Method} & \multirow{2}{*}{Alignment} & \multicolumn{3}{c}{ADT} & \multicolumn{3}{c}{TUM} \\
\cmidrule(lr){3-5}\cmidrule(lr){6-8}
& & PSNR$\uparrow$ & SSIM$\uparrow$ & LPIPS$\downarrow$ & PSNR$\uparrow$ & SSIM$\uparrow$ & LPIPS$\downarrow$ \\
\midrule
\multirow{3}{*}{NeoVerse~\citep{yang2026neoverse}}
& \xmark & 12.217 & 0.2982 & 0.6391 & 14.398 & 0.4159 & 0.4587 \\
& \cmark & 16.179 & 0.4384 & \cellcolor{tabthird}0.5296 & 16.120 & 0.5266 & \cellcolor{tabthird}0.3981 \\
&& {\scriptsize\textcolor{mygreen}{($+$3.962)}}
 & {\scriptsize\textcolor{mygreen}{($+$0.1402)}}
 & {\scriptsize\textcolor{mygreen}{($-$0.1095)}}
 & {\scriptsize\textcolor{mygreen}{($+$1.721)}}
 & {\scriptsize\textcolor{mygreen}{($+$0.1107)}}
 & {\scriptsize\textcolor{mygreen}{($-$0.0606)}} \\
\midrule
\multirow{3}{*}{C4G~\citep{c4g}}
& \xmark & 12.963 & 0.4487 & 0.6870 & 12.378 & 0.3935 & 0.5984 \\
& \cmark & \cellcolor{tabthird}18.900 & \cellcolor{tabthird}0.5813 & 0.5833 & \cellcolor{tabthird}17.508 & \cellcolor{tabthird}0.5518 & 0.4951 \\
&& {\scriptsize\textcolor{mygreen}{($+$5.937)}}
 & {\scriptsize\textcolor{mygreen}{($+$0.1326)}}
 & {\scriptsize\textcolor{mygreen}{($-$0.1038)}}
 & {\scriptsize\textcolor{mygreen}{($+$5.130)}}
 & {\scriptsize\textcolor{mygreen}{($+$0.1583)}}
 & {\scriptsize\textcolor{mygreen}{($-$0.1033)}} \\
\midrule
\multirow{3}{*}{\textbf{\ours{} (Ours)}}
& \xmark & \cellcolor{tabsecond}19.693 & \cellcolor{tabsecond}0.5948 & \cellcolor{tabsecond}0.4490 & \cellcolor{tabsecond}17.847 & \cellcolor{tabsecond}0.5586 & \cellcolor{tabsecond}0.3449 \\
& \cmark & \cellcolor{tabfirst}23.288 & \cellcolor{tabfirst}0.6648 & \cellcolor{tabfirst}0.3683 & \cellcolor{tabfirst}20.392 & \cellcolor{tabfirst}0.6643 & \cellcolor{tabfirst}0.2860 \\
&& {\scriptsize\textcolor{mygreen}{($+$3.594)}}
 & {\scriptsize\textcolor{mygreen}{($+$0.0700)}}
 & {\scriptsize\textcolor{mygreen}{($-$0.0807)}}
 & {\scriptsize\textcolor{mygreen}{($+$2.545)}}
 & {\scriptsize\textcolor{mygreen}{($+$0.1057)}}
 & {\scriptsize\textcolor{mygreen}{($-$0.0589)}} \\
\bottomrule
\end{tabularx}%
\end{table}

%% file: figures/supp_baseline_qual2.tex
\begin{figure}[t]
\centering

\includegraphics[width=.85\textwidth]{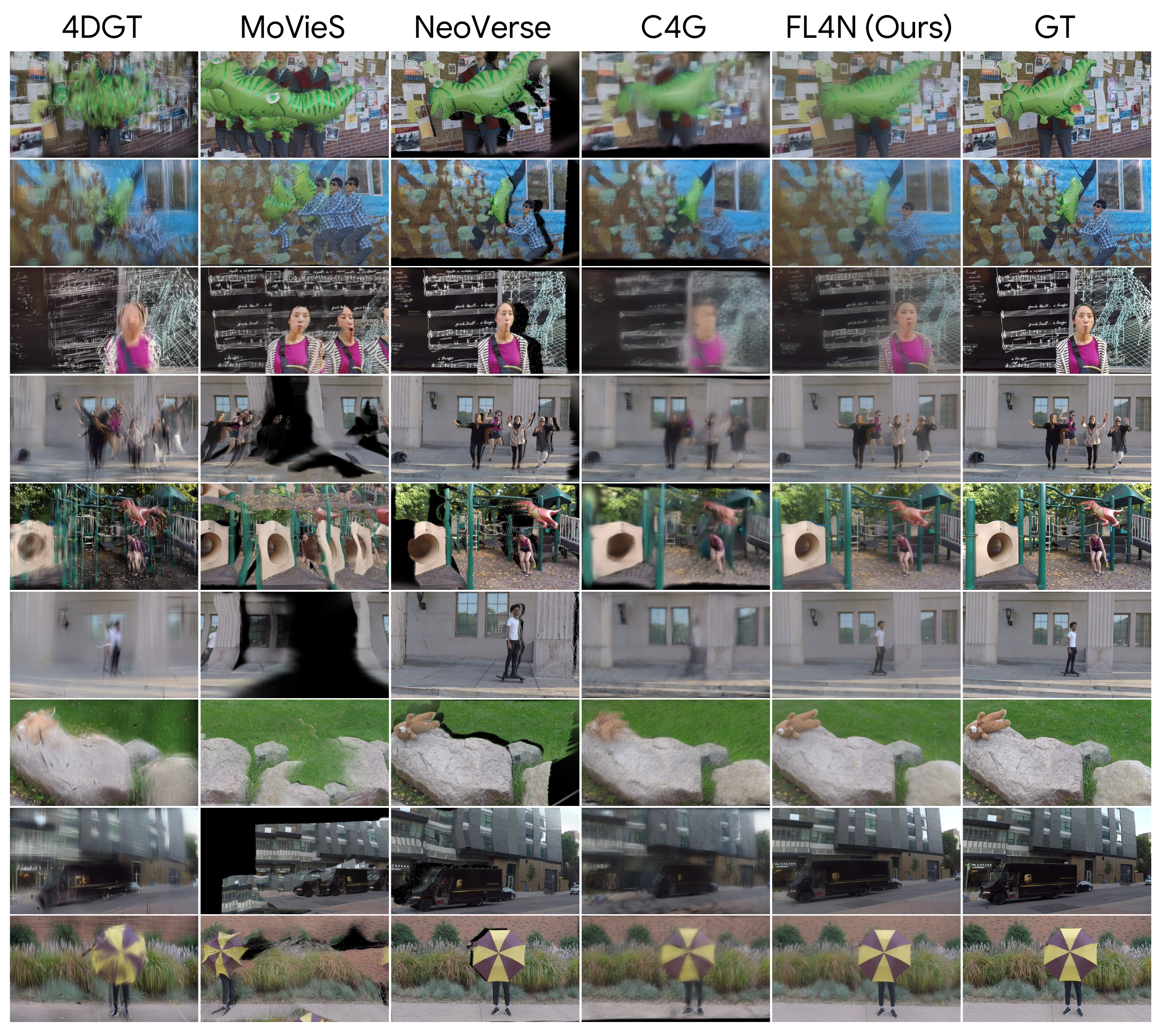}
\vspace{-10pt} 

\caption{\textbf{More qualitative comparisons on NVIDIA across all scenes.} Our method shows much better rendering quality than all other methods based on 3D Gaussian.}
\label{fig:supp_baseline_qual2}
\end{figure}
\vspace{-20pt}

%% file: figures/supp_baseline_qual3.tex
\begin{figure}[t]
\centering

\includegraphics[width=0.88\textwidth]{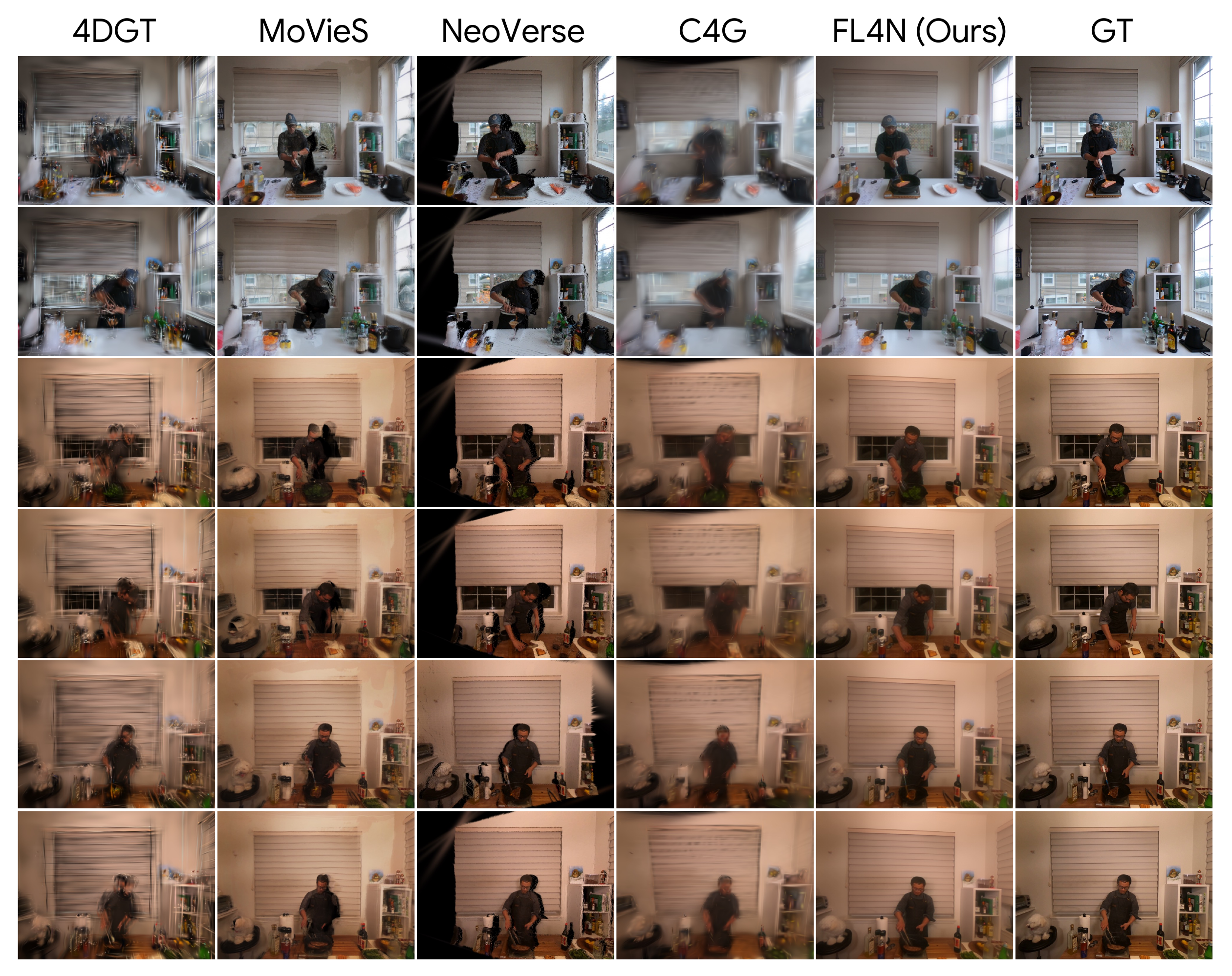}
\vspace{-10pt} 

\caption{\textbf{More qualitative comparisons on N3DV across all scenes.}
Our method shows much better rendering quality than all other methods based on 3D Gaussian.}
\label{fig:supp_baseline_qual3}
\end{figure}

%% file: figures/supp_baseline_qual6.tex
\begin{figure}[t]
\centering

\includegraphics[width=.85\textwidth]{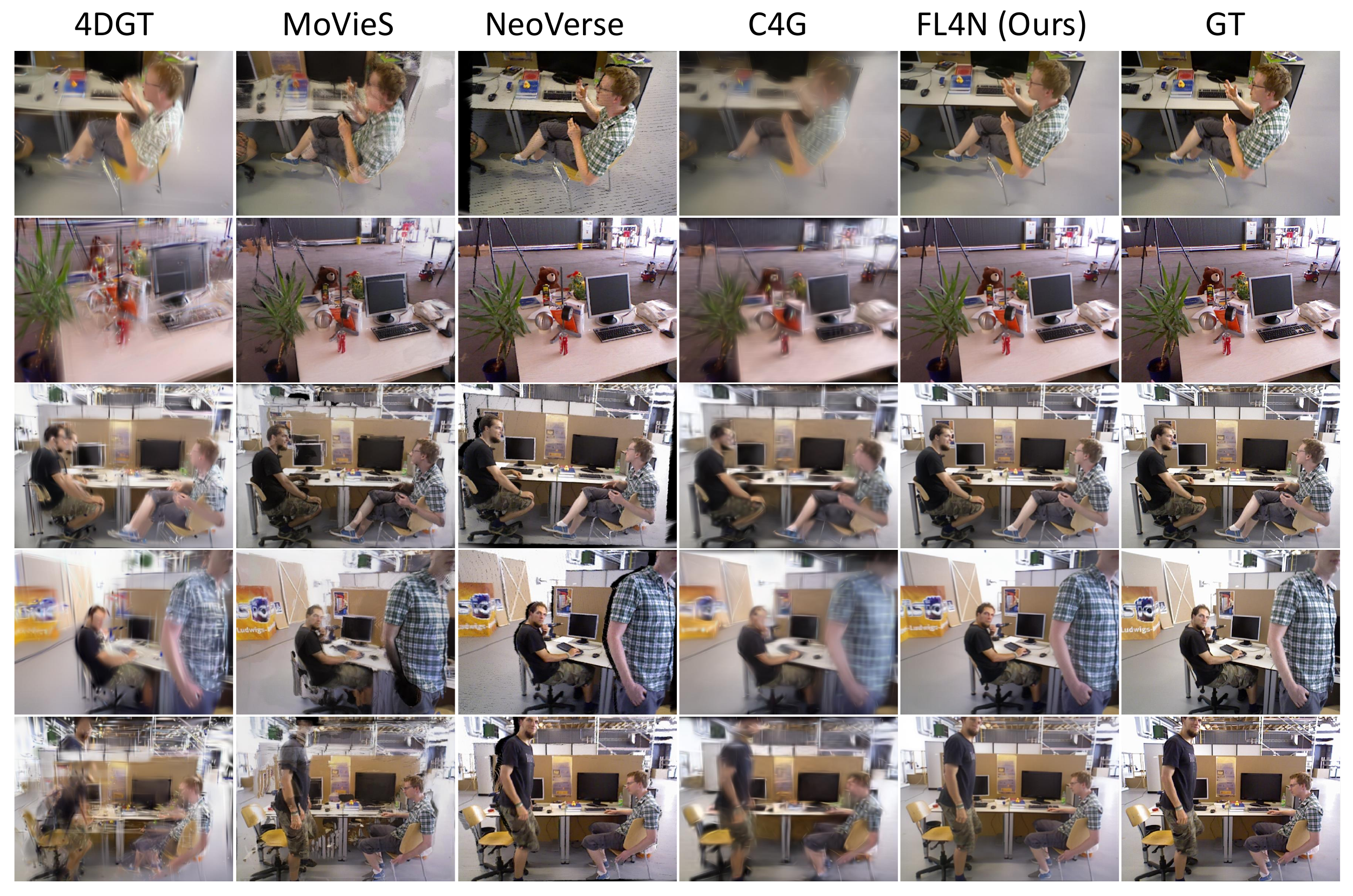}
\vspace{-10pt} 
\caption{\textbf{More qualitative comparisons on TUM across all scenes.}}
\label{fig:supp_baseline_qual6}
\end{figure}
\vspace{-15pt} 

%% file: figures/supp_baseline_qual1.tex
\begin{figure}[t]
\centering

\includegraphics[width=.85\textwidth]{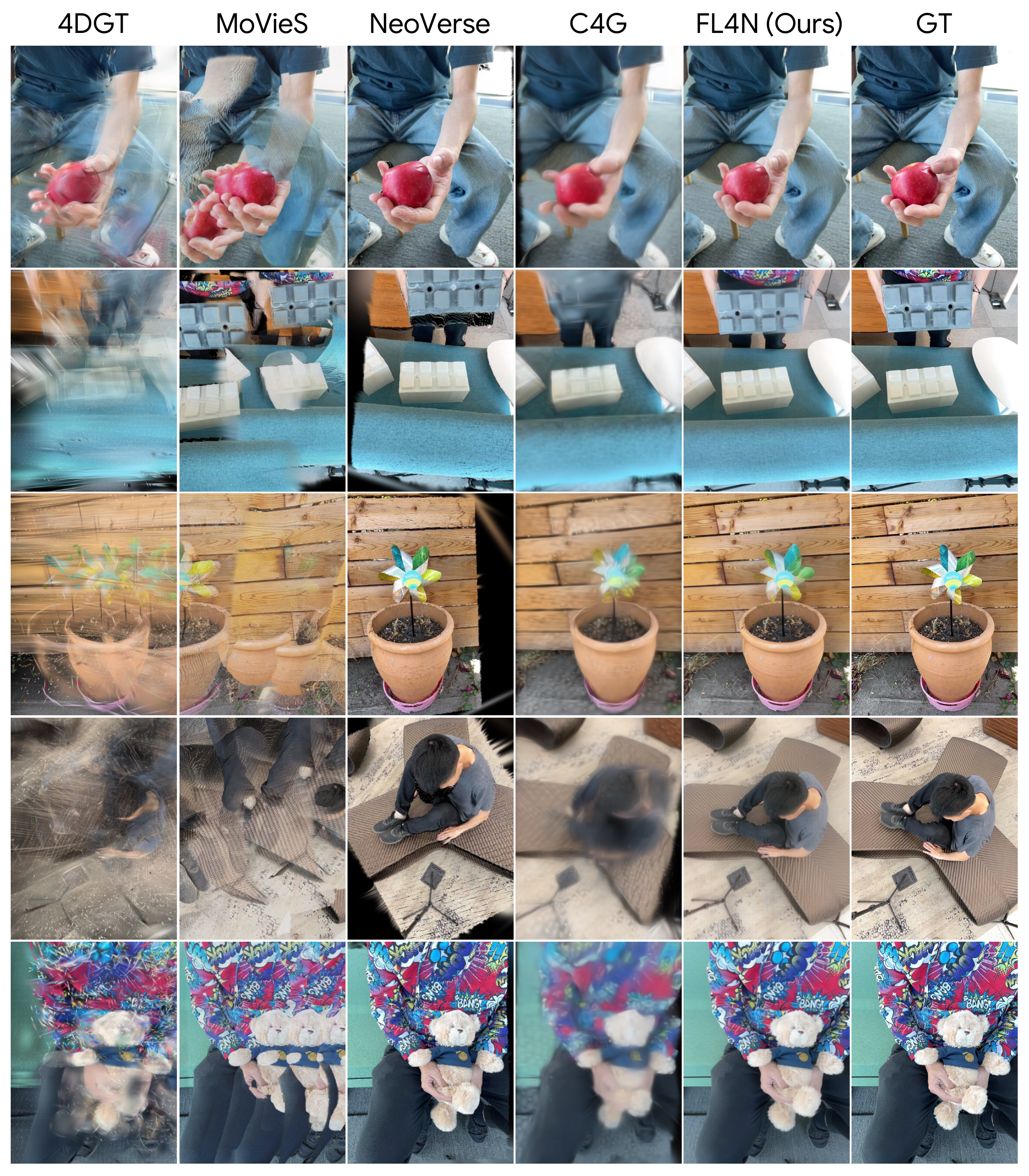}
\vspace{-10pt} 

\caption{\textbf{More qualitative comparisons on DyCheck across all scenes.}}
\label{fig:supp_baseline_qual1}
\end{figure}

%% file: figures/supp_baseline_qual4.tex
\begin{figure}[t]
\centering

\includegraphics[width=\textwidth]{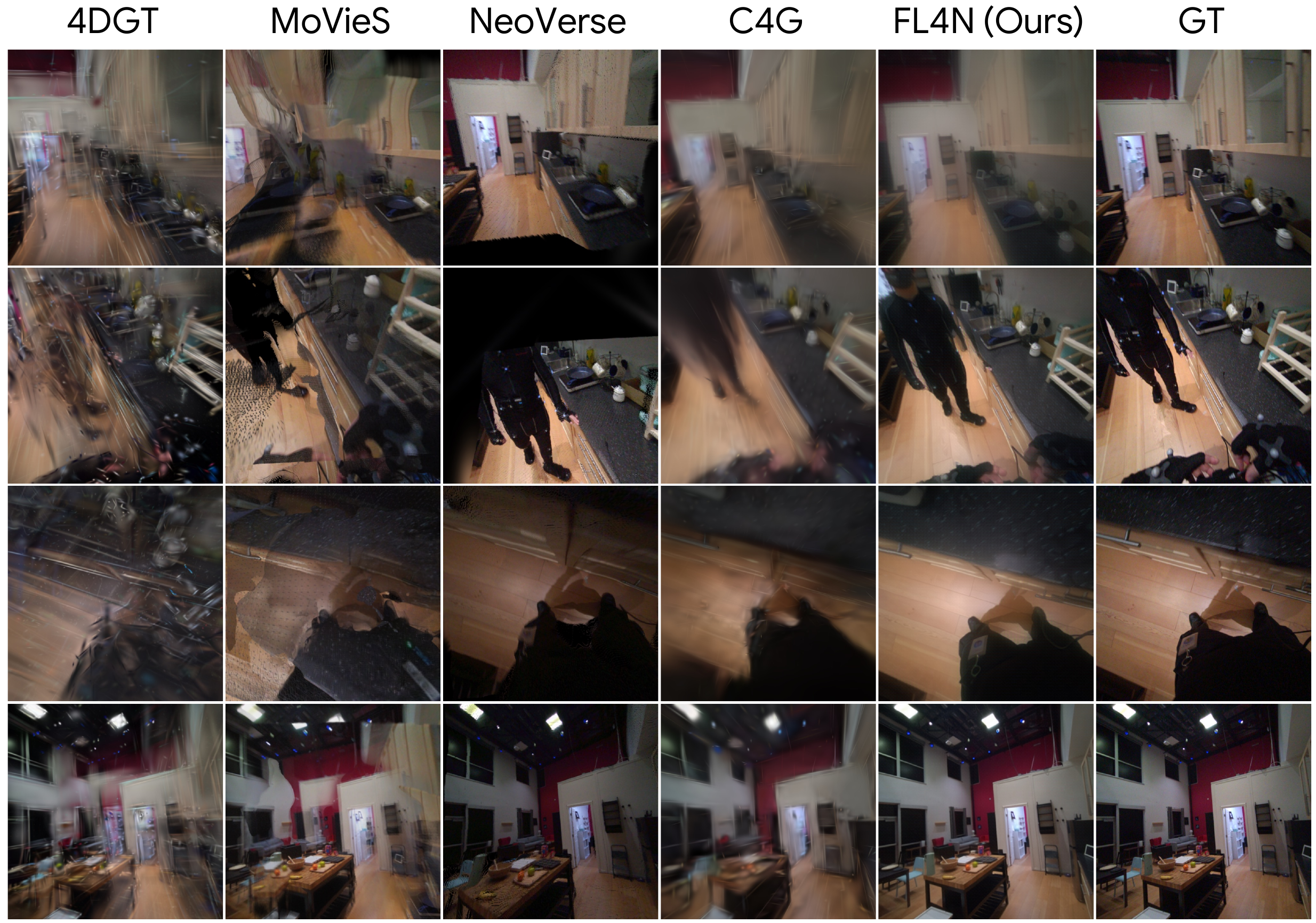}
\vspace{-5pt} 
\caption{\textbf{More qualitative comparisons on ADT across all scenes.}
Our method shows much better rendering quality than all other methods based on 3D Gaussian.
Gaussian-based approaches show severe artifacts such as holes caused by misaligned/mispositioned Gaussians.}
\label{fig:supp_baseline_qual4}
\end{figure}

%% file: figures/supp_baseline_qual5.tex
\begin{figure}[t]
\centering

\includegraphics[width=\textwidth]{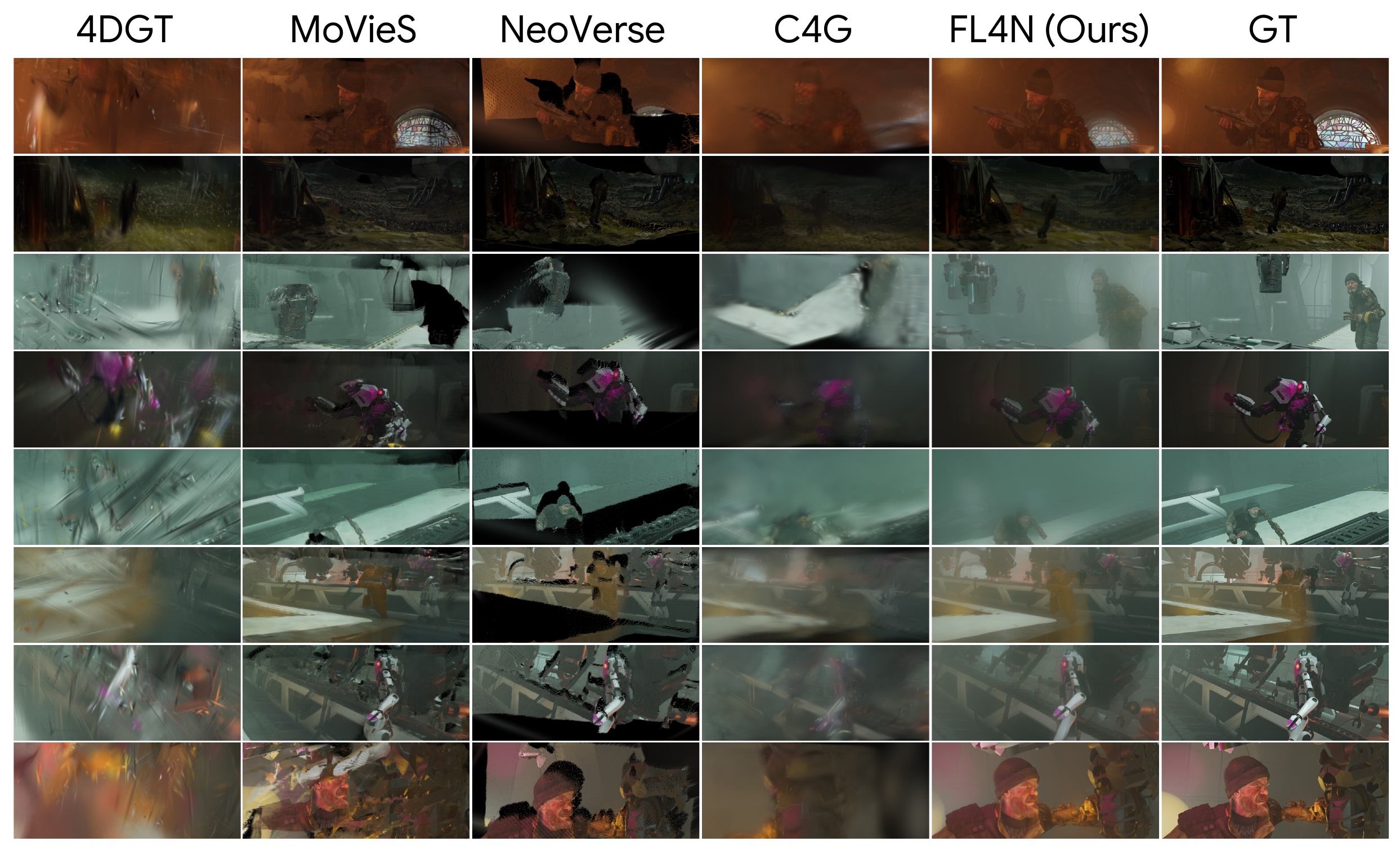}
\vspace{-9pt} 
\caption{\textbf{More qualitative comparisons on Charge across all scenes.}
Our method shows much better rendering quality than all other methods based on 3D Gaussian.
Gaussian-based approaches show severe artifacts such as holes caused by misaligned/mispositioned Gaussians.
}
\label{fig:supp_baseline_qual5}
\end{figure}

%% file: figures/supp_baseline_ctvs_qual_1.tex
\begin{figure}[t]
\centering

\includegraphics[width=\textwidth]{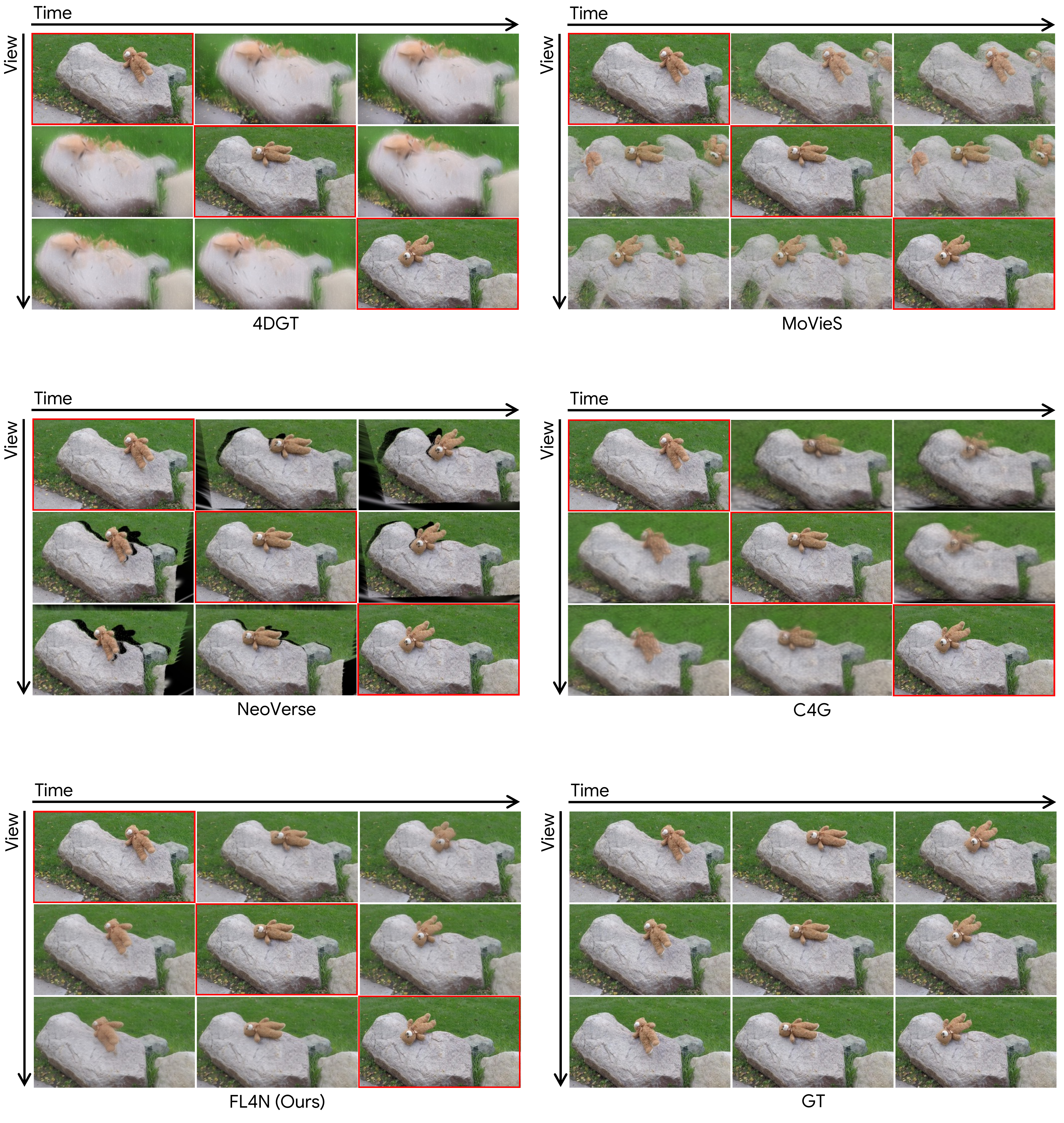}
\vspace{-10pt} 

\caption{\textbf{Qualitative comparisons across view-time grid on NVIDIA.} Red outlines indicate context frames.
Compared with other methods, our methods show substantially better rendering quality with clear view-time disentanglement.}
\label{fig:supp_baseline_ctvs_qual1}
\end{figure}

%% file: figures/supp_baseline_ctvs_qual_3.tex
\begin{figure}[t]
\centering

\includegraphics[width=\textwidth]{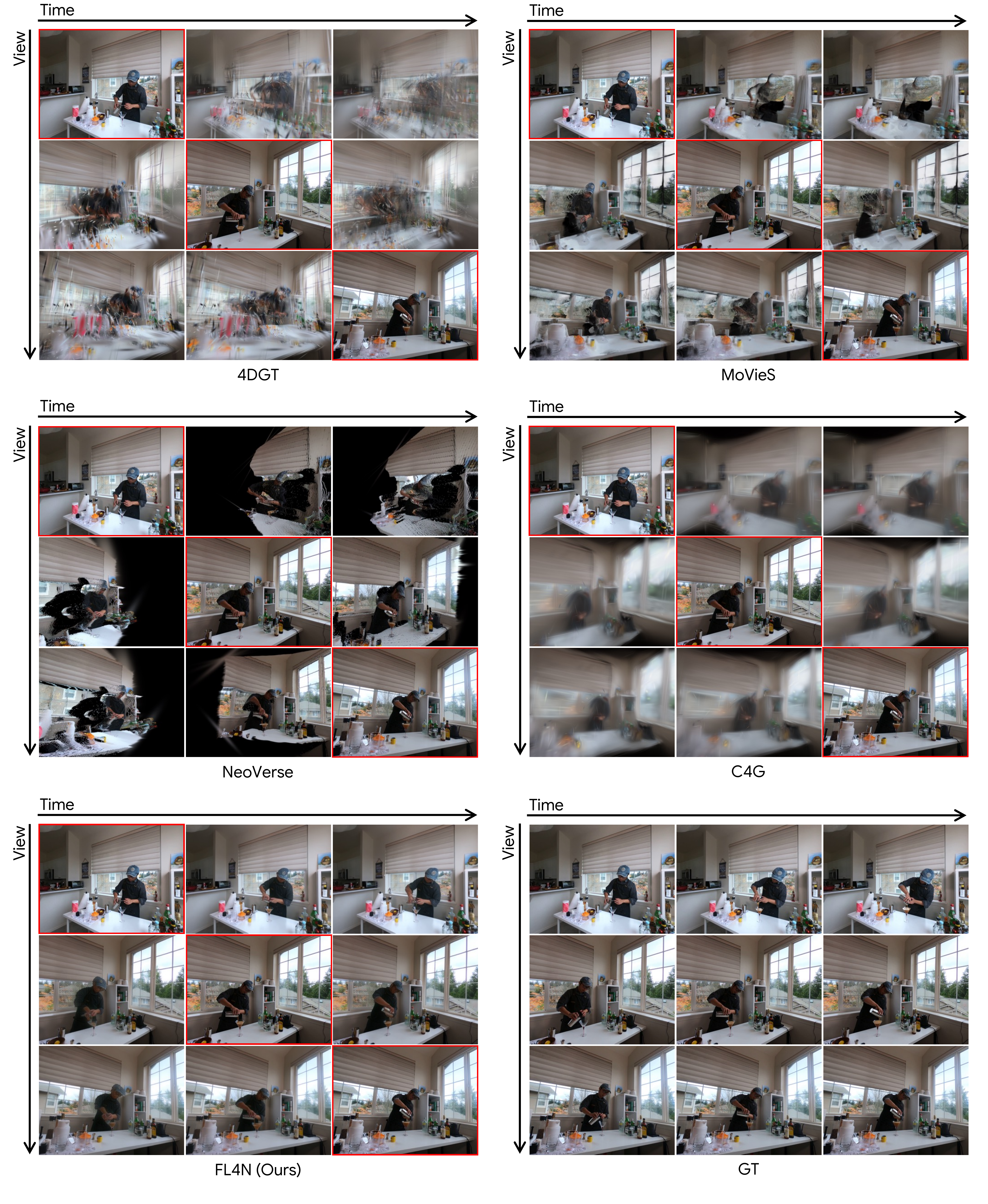}
\vspace{-10pt}

\caption{\textbf{Qualitative comparisons across view-time grid on N3DV.} Red outlines indicate context frames.
Compared with other methods, our methods show substantially better rendering quality with clear view-time disentanglement.}
\label{fig:supp_baseline_ctvs_qual3}
\end{figure}

%% file: figures/supp_baseline_ctvs_qual_2.tex
\begin{figure}[t]
\centering

\includegraphics[width=1.0\textwidth]{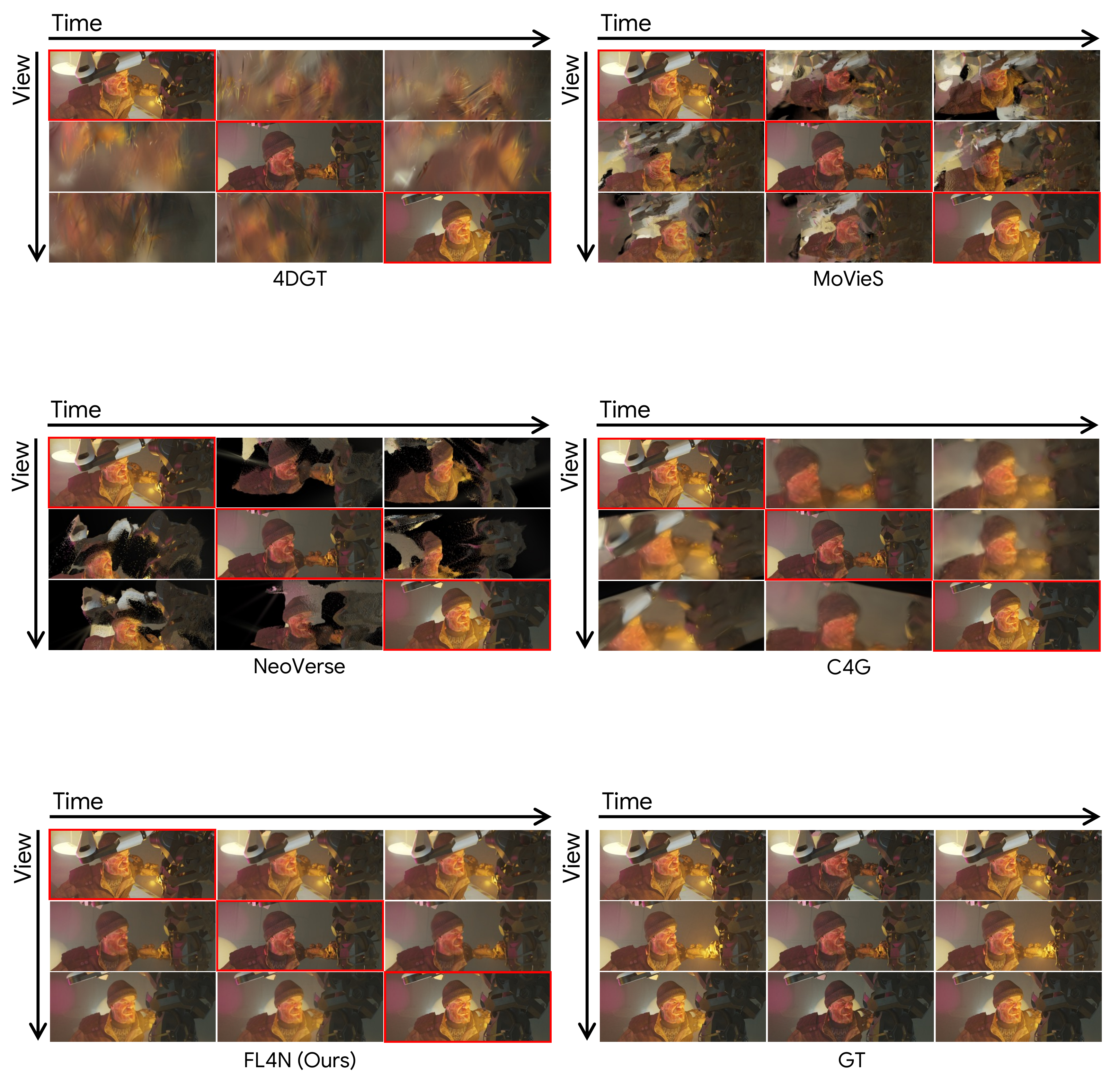}
\vspace{-10pt}

\caption{\textbf{Qualitative comparisons across view-time grid on Charge.} Red outlines indicate context frames.
Compared with other methods, our methods show substantially better rendering quality with clear view-time disentanglement.}
\label{fig:supp_baseline_ctvs_qual2}
\end{figure}

%% file: figures/supp_nvs_qual1.tex
\begin{figure}[t]
\centering

\includegraphics[width=.9\textwidth]{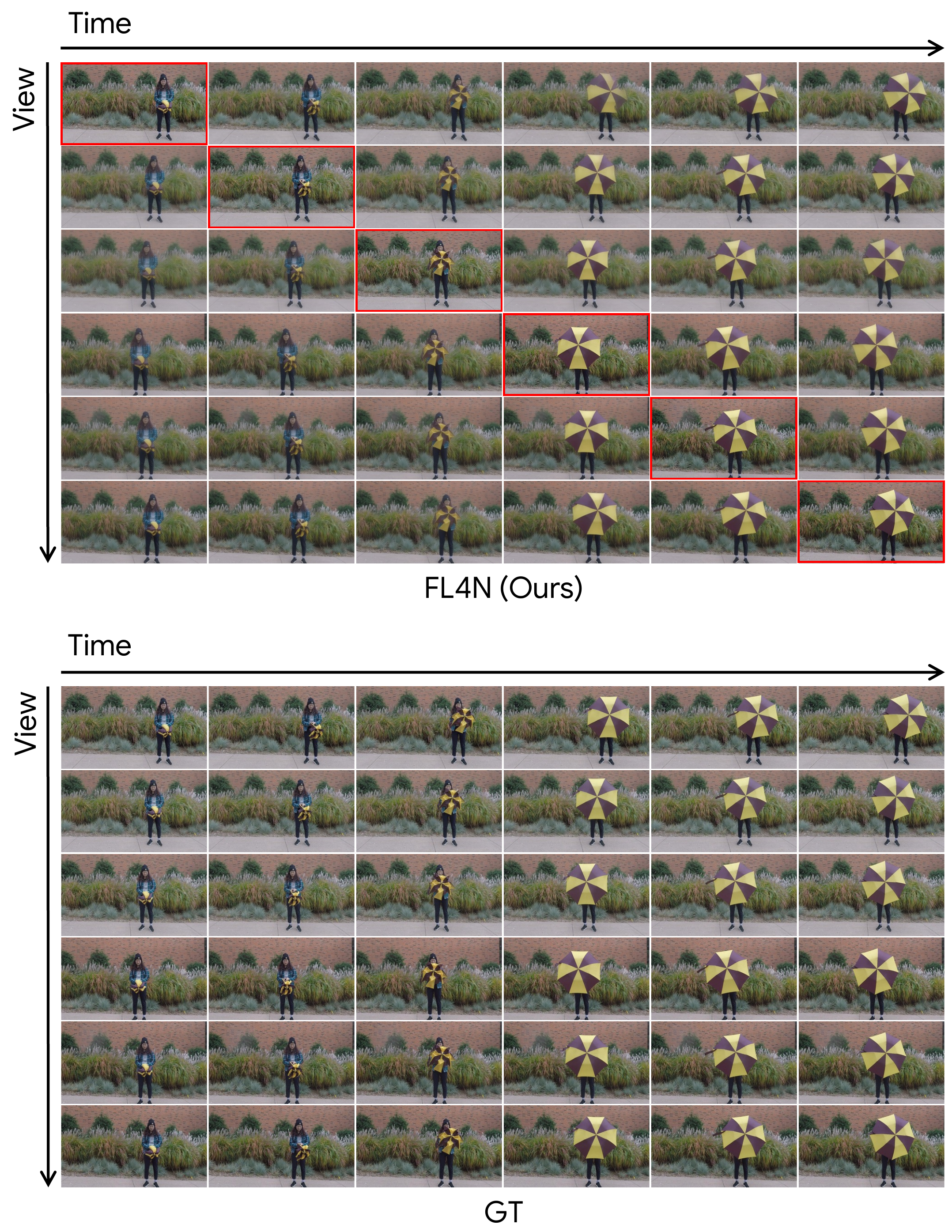}

\caption{\textbf{Qualitative results across view-time grid on NVIDIA.} Red outlines indicate context frames.
Our method successfully synthesizes images at novel view and time, using target conditioning on predicted latent, without relying on explicit geometric primitives.
}
\label{fig:supp_nvsqual1}
\end{figure}

%% file: figures/supp_nvs_qual2.tex
\begin{figure}[t]
\centering

\includegraphics[width=0.7\textwidth]{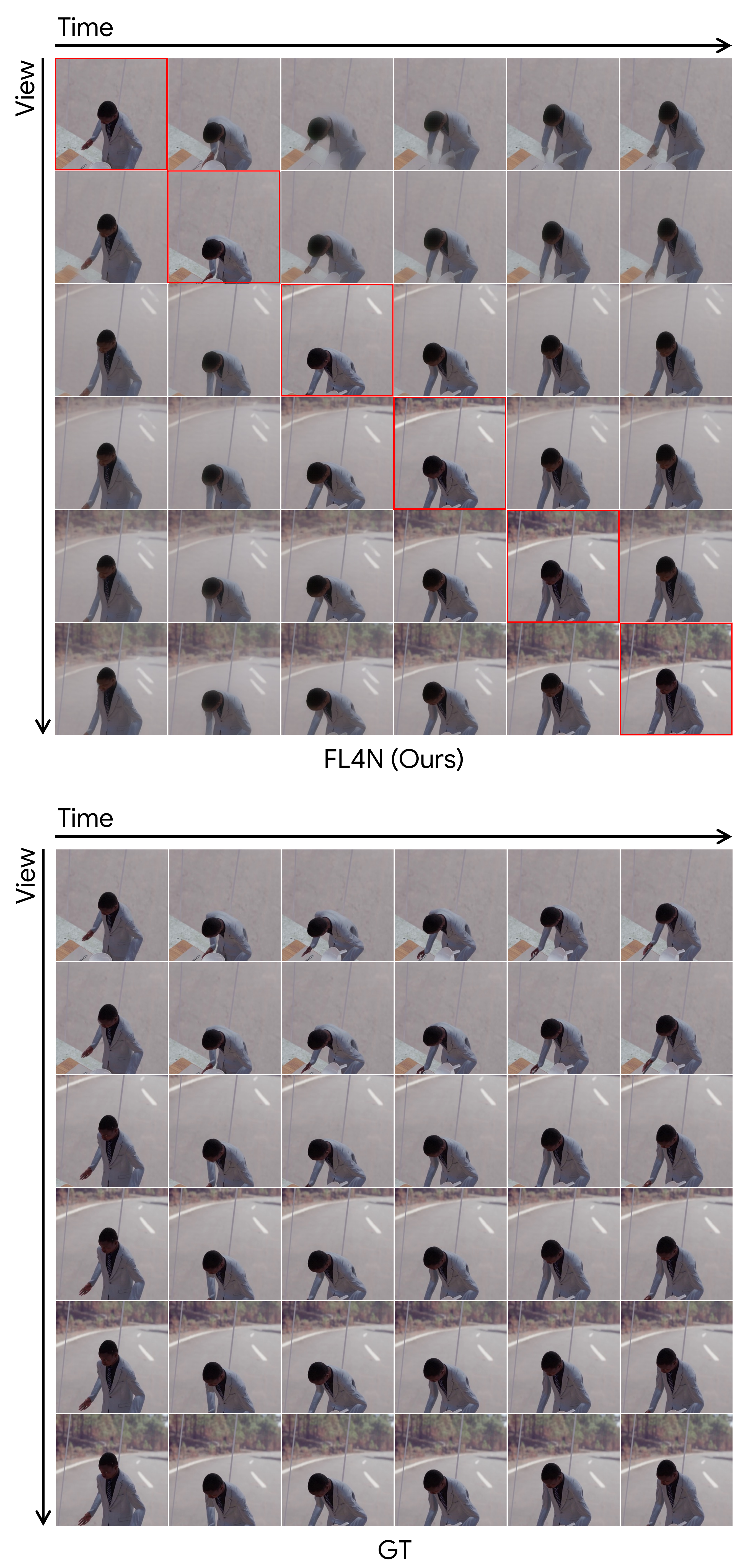}

\caption{\textbf{Qualitative results across view-time grid on CamxTime.}
Red outlines indicate context frames.
Our method successfully synthesizes images at novel view and time, using target conditioning on predicted latent, without relying on explicit geometric primitives.
}
\label{fig:supp_nvsqual2}
\end{figure}

%% file: figures/supp_nvs_qual3.tex
\begin{figure}[t]
\centering

\includegraphics[width=0.9\textwidth]{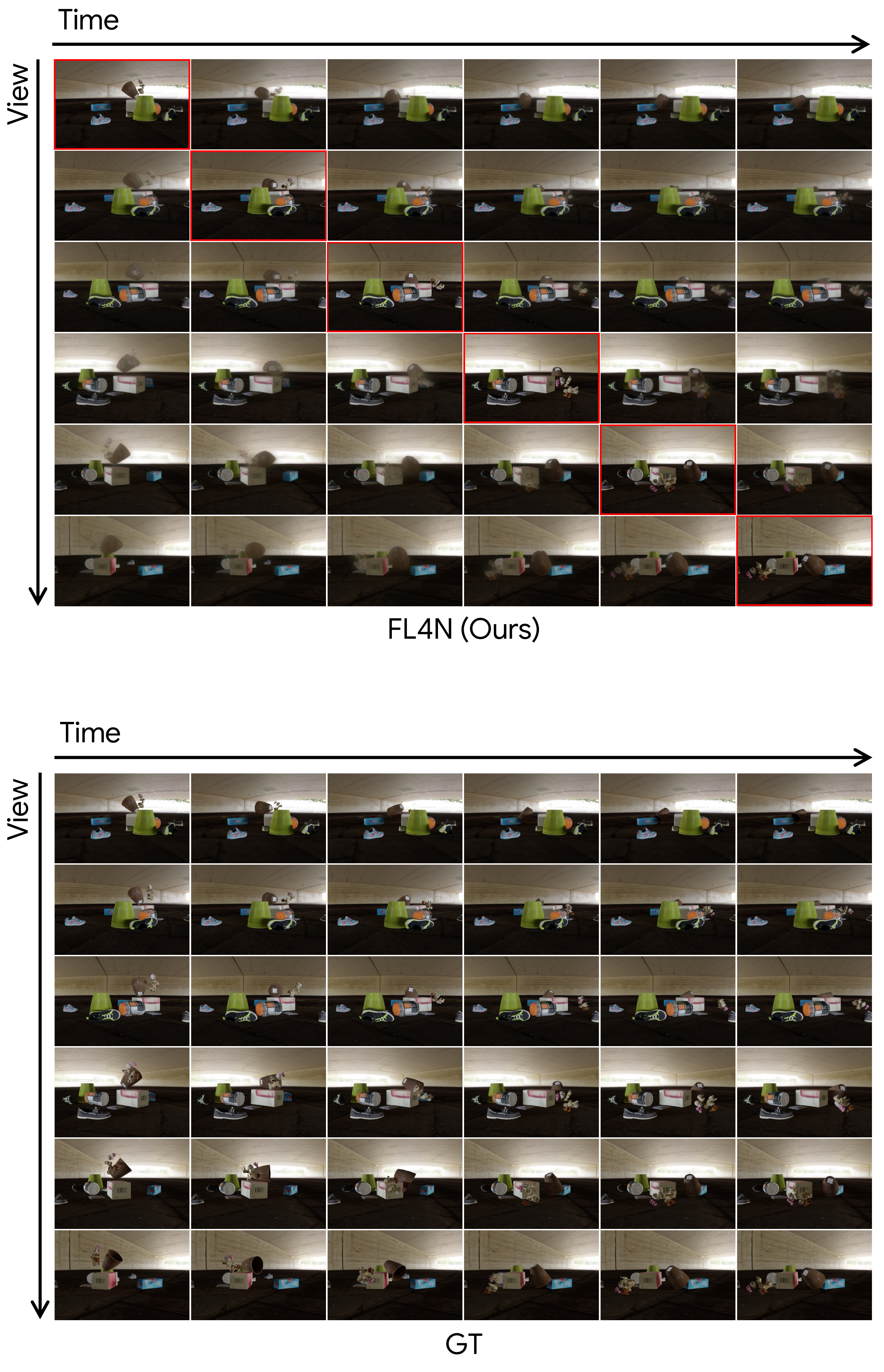}

\caption{\textbf{Qualitative results across view-time grid on Kubric4D.} Red outlines indicate context frames.
Our method successfully synthesizes images at novel view and time, using target conditioning on predicted latent, without relying on explicit geometric primitives.
}
\label{fig:supp_nvsqual3}
\end{figure}